\documentclass[11pt]{article}

\usepackage[a4paper,margin=1in]{geometry}
\usepackage{iftex}
\ifPDFTeX
  \usepackage[T1]{fontenc}
  \usepackage[utf8]{inputenc}
  \usepackage{lmodern}
  \DeclareUnicodeCharacter{00D1}{\~{N}}
  \DeclareUnicodeCharacter{00F1}{\~{n}}
  \DeclareUnicodeCharacter{00E0}{\`{a}}
  \DeclareUnicodeCharacter{00E1}{\'{a}}
  \DeclareUnicodeCharacter{00E9}{\'{e}}
  \DeclareUnicodeCharacter{0107}{\'{c}}
  \DeclareUnicodeCharacter{0100}{\={A}}
  \DeclareUnicodeCharacter{0101}{\={a}}
  \DeclareUnicodeCharacter{012B}{\={\i}}
  \DeclareUnicodeCharacter{016B}{\={u}}
  \DeclareUnicodeCharacter{1E41}{\.{m}}
  \DeclareUnicodeCharacter{1E43}{\d{m}}
  \DeclareUnicodeCharacter{1E45}{\.{n}}
  \DeclareUnicodeCharacter{1E47}{\d{n}}
  \DeclareUnicodeCharacter{1E6D}{\d{t}}
  \DeclareUnicodeCharacter{2013}{--}
  \DeclareUnicodeCharacter{2026}{\ldots{}}
  \DeclareUnicodeCharacter{2192}{$\rightarrow$}
\else
  \usepackage{fontspec}
\fi

\usepackage{microtype}
\usepackage{booktabs}
\usepackage{array}
\usepackage{xltabular}
\usepackage{ragged2e}
\usepackage{enumitem}
\usepackage{fvextra}
\usepackage{xurl}
\usepackage{caption}
\usepackage[hidelinks]{hyperref}

\setlist[itemize]{leftmargin=1.5em,itemsep=0.15em,topsep=0.25em}
\DefineVerbatimEnvironment{CodeBlock}{Verbatim}{
  breaklines,
  breakanywhere,
  fontsize=\small,
  frame=single,
  framesep=2mm,
  xleftmargin=0.5em,
  xrightmargin=0.5em
}

\newcolumntype{Y}{>{\RaggedRight\arraybackslash}X}

\title{From Outliers to Errors: Auditing Pali-to-English LLM Translations with Multi-Reference Adjudication}
\author{Máté Metzger, Independent Researcher, Hungary \and Nadnapang Phophichit, Nibbana Meditation Centre, Thailand \and Hansa Dhammahaso, Nibbana Meditation Centre, Thailand}
\date{}
\begin{document}
\maketitle
\begin{abstract}
Single-score translation metrics can conflate legitimate variation with error, a problem especially acute for classical languages where multiple defensible English renderings of the same passage coexist. We audit Pali-to-English output from four flagship large language models (LLMs)--GPT-5.5, Claude Sonnet 4.6, Gemini 3.1 Pro, and Grok 4.3--on 1,700 passages from the Pali Canon, using three established human translations by Bhikkhu Sujato, Thanissaro Bhikkhu, and Bhikkhu Bodhi as a local reference envelope rather than a single gold standard. Each candidate's normalized embedding drift from the reference centroid serves as a triage signal, not an error label; the 1,203 candidates above a \texttt{1.\allowbreak{}5} drift threshold are then adjudicated by a blinded three-model LLM judge panel, calibrated against a 300-instance author-adjudicated validation set. Two results stand out. First, drift predicts severity rather than error per se: the major-error rate among adjudicated high-drift candidates rose monotonically from 7.9\% in the \texttt{1.\allowbreak{}5-\allowbreak{}2.\allowbreak{}0} band to 51.6\% above \texttt{3.\allowbreak{}0}, while approximately 80\% of \texttt{1.\allowbreak{}5-\allowbreak{}2.\allowbreak{}0} outliers were judged valid translation variations. Second, model differences were clearest in the high-drift tail: GPT-5.5 had the lowest adjudicated high-drift major-error rate, with confidence intervals overlapping those of Claude Sonnet 4.6 and Gemini 3.1 Pro; Grok 4.3 had both the largest outlier volume and the highest tail major-error rate (27.6\% overall, 74.4\% above drift \texttt{3.\allowbreak{}0}). The dominant major-error categories--omission or truncation, doctrinal term errors, agent or role errors, additions or hallucinations, and negation or polarity errors--are precisely the failures most likely to mislead readers of doctrinal text. The contribution is a reusable audit design for classical-to-modern translation: define a local reference envelope from multiple human translators, use embedding drift to prioritize review, and adjudicate the flagged tail with calibrated severity labels rather than treating outlier status as error.
\end{abstract}
\section{Introduction}

Large language models (LLMs) are increasingly plausible translators for texts and language pairs that sit outside the center of standard machine-translation benchmarks. Recent studies find that Generative Pre-trained Transformer (GPT)-class systems can perform competitively on some translation directions, while still showing strong variation by language, domain, prompt, and resource level [Jiao et al. 2023]; [Hendy et al. 2023]; [Zhu et al. 2024]; [Xu et al. 2024]. At the same time LLMs are evolving rapidly. This makes translation of classical and low-resource languages a particularly important test case. These languages often have limited parallel data, specialized terminology, historical registers, and interpretive traditions that are poorly represented by modern web text.

Pali is a Middle Indo-Aryan language best known as the language of the Pali Canon, the scriptural corpus of Theravada Buddhism. Pali canonical texts are formulaic in some places, dense and elliptical in others, and often depend on a technical Buddhist vocabulary whose English rendering is both consequential and legitimately variable. Human translators differ in style, compression, terminology, and interpretive stance. A translation can diverge from one reference and still be defensible; conversely, a fluent and reference-like translation can contain a substantive error.

Part of this problem can be addressed by using multiple independent translations as references. In this study, the selected references are not interchangeable targets but works by translators with distinct styles, audiences, and interpretive commitments: Bhikkhu Sujato's plain and accessible digital reading environment on SuttaCentral, Thanissaro Bhikkhu's practice-oriented corpus, and Bhikkhu Bodhi's scholarly editions. Since translation is an interpretive and culturally situated act rather than a transparent transfer of meaning [Venuti 2018]; [Tymoczko 2014], the three references are treated here as interpretive witnesses that help define a range of defensible renderings, not as a neutral oracle.

Multi-reference evaluation is especially relevant for texts where translation variation is not noise but evidence of the range of acceptable renderings [Fomicheva, Specia, and Guzmán 2020]; [Wu, Wieting, and Smith 2024]. However, multi-reference metrics still do not answer a central scholarly question: when an LLM translation falls outside the human-reference envelope, is it wrong, or merely different? This paper treats that question as the object of study.

The main contribution is a two-stage audit framework. First, candidate translations are embedded and compared against the local distribution of human references. This produces a normalized drift score that functions as a triage signal. Second, high-drift candidates are adjudicated with explicit labels for valid variation, minor error, and major error. The distinction matters. Automatic metrics such as Bilingual Evaluation Understudy (BLEU), character n-gram F-score (chrF), learned neural metrics, and embedding similarity are useful summaries, but prior work has repeatedly warned against treating scalar scores as direct substitutes for human error analysis [Papineni et al. 2002]; [Popović 2015]; [Rei et al. 2020]; [Freitag et al. 2022]; [Mathur et al. 2020]. Our design therefore asks not only which model is closer to references on average, but which outputs contain correction-worthy or materially misleading translation errors, and which are valid renderings despite differing from the references.

The study evaluates four flagship translator models on 1,700 selected Pali passages and adjudicates the high-drift portion of their outputs with a calibrated LLM judge panel. The goal is not to claim that LLM judges replace expert philological assessment. Instead, the goal is to test whether a scalable, reference-aware audit pipeline can separate legitimate variation from likely errors well enough to estimate error rates, error types, and model-specific failure patterns. The resulting blueprint is intended to generalize to other classical-to-modern language settings where multiple human references exist but exhaustive expert review is impractical.

The paper is organized around four research questions. First, how strongly does multi-reference embedding drift predict minor and major translation errors? Second, among high-drift translations, what proportion are genuine errors rather than valid variations? Third, what types of Pali-to-English errors do flagship LLMs make, and do those types differ across models? Fourth, do serious failures cluster on the same difficult source passages, or distributed more broadly? These questions shift the evaluation target from model ranking alone to error discovery, severity estimation, and reusable audit design.

\section{Literature Review}

Machine-translation evaluation has long moved between automatic scoring and human error analysis. BLEU established the dominant early paradigm of corpus-level n-gram overlap [Papineni et al. 2002], while later lexical and character-level metrics such as Metric for Evaluation of Translation with Explicit ORdering (METEOR) and chrF addressed some of BLEU's brittleness, especially around morphology and partial matching [Lavie and Denkowski 2009]; [Popović 2015]; [Popović 2017]. Subsequent work pushed evaluation toward contextual and learned semantic metrics, including BERTScore, Bilingual Evaluation Understudy with Representations from Transformers (BLEURT), and Crosslingual Optimized Metric for Evaluation of Translation (COMET) [Zhang et al. 2020]; [Sellam, Das, and Parikh 2020]; [Rei et al. 2020]. The Workshop on Machine Translation (WMT) metrics literature suggests that neural metrics often correlate better with human judgments than BLEU, but also stresses that metric choice and reporting details can change conclusions [Post 2018]; [Freitag et al. 2022]; [Mathur et al. 2020].

This paper builds on that literature but uses automatic metrics differently. Rather than using a scalar metric as the final outcome, we use multi-reference embedding distance as an audit filter. This choice follows two observations. First, multiple references better represent acceptable translation variation than a single reference, particularly for literary or interpretive texts [Fomicheva, Specia, and Guzmán 2020]; [Wu, Wieting, and Smith 2024]. Second, embedding models can capture semantic proximity beyond surface overlap, but they are not neutral arbiters of translation correctness. Sentence and multilingual embedding work has made semantic comparison more practical [Reimers and Gurevych 2019]; [Artetxe and Schwenk 2019]; [Feng et al. 2022], while embedding benchmark studies caution that no embedding method dominates every task [Muennighoff et al. 2023]. In this study, embedding drift is therefore a way to find suspicious cases, not a way to declare them erroneous.

The adjudication layer draws on translation quality assessment and error typology traditions. Multidimensional Quality Metrics (MQM)-style evaluation emphasizes explicit error categories, severity, and expert judgment rather than undifferentiated adequacy scores [Lommel 2018]; [MQM Council n.d.]. Large-scale human machine-translation (MT) evaluation has shown that expert error annotation with context can expose quality differences that simpler ratings or automatic metrics obscure [Freitag et al. 2021]. Error-focused automatic metrics such as xCOMET similarly reflect a broader movement from global scores toward transparent error detection [Guerreiro et al. 2024]. Our labels are intentionally coarser than full MQM: they separate valid variation, minor error, and major error, then assign a limited set of categories designed for Pali-to-English translation. This narrower taxonomy is meant to support reliable adjudication at scale while preserving the distinction most important for downstream users: whether a reader would be materially misled.

Related literature also discusses hallucination and severe translation failures. Early MT error analysis already argued that aggregate metrics do not identify the main sources of system failure [Vilar et al. 2006]. Neural MT and LLM translation add renewed concern about hallucinated, omitted, or semantically unrelated output [Raunak, Menezes, and Junczys-Dowmunt 2021]; [Guerreiro et al. 2023a]; [Guerreiro et al. 2023b]; [Dale et al. 2023]; [Benkirane et al. 2024]. These studies are directly relevant to our distinction between minor and major errors. A minor terminology preference may be correction-worthy but not misleading; however, an omission, polarity reversal, wrong agent, or hallucinated addition can change the reader's understanding of the passage. The present study asks whether such serious failures cluster in the high-drift tail and whether their categories differ by model.

LLM-based judging is the other methodological pillar. GPT Estimation Metric Based Assessment (GEMBA) and GEMBA-MQM showed that LLMs can be prompted to score MT quality and identify MQM-like error spans [Kocmi and Federmann 2023a]; [Kocmi and Federmann 2023b]. Work such as G-Eval and MT-Bench/Chatbot Arena further demonstrates the utility and risks of LLM-as-judge designs for language-generation evaluation [Liu et al. 2023]; [Zheng et al. 2023]; [Chiang et al. 2024]. The risk is particularly acute when the judge is asked to adjudicate subtle translation variation in a specialized domain. For that reason, our LLM panel is independent of the translator panel, judges are blinded to translator identity, and a 300-instance author-adjudicated validation set is used to measure calibration before interpreting the full result set.

Finally, the study is situated within computational work on ancient and low-resource languages. No Language Left Behind (NLLB) and FLORES Evaluation Benchmark for Low-Resource and Multilingual Translation (FLORES) demonstrate the scale and importance of multilingual translation benchmarks, but classical languages remain underrepresented in mainstream MT evaluation [Costa-jussà et al. 2022]; [Costa-jussà et al. 2024]; [Goyal et al. 2022]. Ancient-language natural language processing (NLP) has developed specialized models and tasks for Greek, Latin, Sanskrit, and Buddhist textual traditions, including Sanskrit morphological disambiguation and unified Sanskrit NLP models [Sommerschield et al. 2023]; [Assael et al. 2022]; [Bamman and Burns 2020]; [Riemenschneider and Frank 2023]; [Hellwig 2015]; [Nehrdich, Hellwig, and Keutzer 2024]; [Nehrdich and Keutzer 2026]. Recent work on Ancient Chinese and Ancient Greek translation suggests that modern LLM competence on contemporary languages does not automatically transfer to classical texts, and that rare technical terminology can predict catastrophic failure [Zhang and Li 2023]; [Zainaldin et al. 2026]. This study contributes to that line of work by focusing not merely on average translation quality, but on the rate, severity, and typology of errors that matter for scholarly use.

The project also belongs to Buddhist digital humanities. Recent work in this area has shown how language-model-based workflows can support distant reading of canonical Buddhist corpora when paired with domain interpretation, as in Bingenheimer, Brody, and Nichols's topic-modeling study of Chinese Buddhist texts [Bingenheimer, Brody, and Nichols 2025]. The present study applies a similar principle to translation evaluation: computational methods are used to prioritize and structure attention, while interpretation remains anchored in philological judgment.

Classical-language translation is also not only a low-resource problem. It is a philological evaluation and a philosophical interpretation problem. The target output must be judged against grammar, inherited commentary and terminology, genre conventions, and the range of defensible translator choices. For Buddhist canonical translation, a rendering that is fluent and semantically adjacent may still be unacceptable if it reverses agency, erases a doctrinal distinction, mishandles a stock formula, or turns an elliptical repetition into an invented assertion. Buddhist technical vocabulary is especially sensitive because apparently small lexical choices can locate a passage within a wider conceptual network [Buswell and Lopez 2014]. This is why the present study combines multi-reference semantic triage with severity-based adjudication rather than relying on a single metric family.

\section{Methods}

\subsection{Dataset}

The source corpus was drawn from Sutta Piṭaka texts of the Pali Canon in the Aṅguttara, Majjhima, and Saṃyutta Nikāyas for which overlapping English translations by three reference translators were publicly available online under licenses or terms permitting scholarly use: SuttaCentral for Sujato and Bodhi [SuttaCentral n.d.] and dhammatalks.org for Thanissaro [Thanissaro Bhikkhu n.d.]. This initial corpus contained 380 SuttaCentral source files (texts): 100 Aṅguttara Nikāya (AN) files, 39 Majjhima Nikāya (MN) files, and 241 Saṃyutta Nikāya (SN) files. In SuttaCentral's data model, a segment is the smallest addressable text unit, identified by a stable JSON key such as \texttt{mn2:\allowbreak{}1.\allowbreak{}1}; a passage is a group of segments sharing the same structural prefix, such as \texttt{mn2:\allowbreak{}1}. These files contained 22,384 SuttaCentral segments grouped into 4,003 passage candidates. Passages are used here as the unit for both translation and evaluation because they are more semantically complete and meaningful than individual segments.

Sujato's Pali and English texts were already available in segmented JavaScript Object Notation (JSON) format with stable segment identifiers from SuttaCentral's GitHub repository. Thanissaro's and Bodhi's independently segmented translations were aligned to those identifiers with GPT-5-mini using structured JSON output: for each batch of 15-30 segments, the model received the Pali source, Sujato's English translation as a segmentation guide, and the target translator's English text, then extracted the minimal corresponding target-translation span for each segment. The alignment system prompt is reproduced in Appendix C. Each extracted Thanissaro and Bodhi segment was then verified against the source files by exact matching, normalized matching, ellipsis-expansion matching, or cross-reference matching.

Filtering was designed to remove incomplete, anomalously aligned, extremely short, or highly repetitive candidates before model evaluation, so that later outlier analysis would reflect translation behavior rather than corpus-construction artifacts. Table 1 summarizes the filtering rules and their rationale.

\par\medskip\noindent\textbf{Table 1. Dataset filtering rules.}\par\smallskip
\begingroup
\small
\setlength{\tabcolsep}{3pt}
\begin{xltabular}{\textwidth}{@{}>{\RaggedRight\arraybackslash}p{0.22\textwidth}Y Y@{}}
\toprule
\textbf{Filter} & \textbf{Rule} & \textbf{Rationale} \\
\midrule
\endfirsthead
\toprule
\textbf{Filter} & \textbf{Rule} & \textbf{Rationale} \\
\midrule
\endhead
\midrule
\multicolumn{3}{r}{\footnotesize continued on next page}\\
\endfoot
\bottomrule
\endlastfoot
Missing text & Exclude if any constituent segment had a null text field. & Remove incomplete source or reference items. \\
Failed verification & Exclude if any extracted segment failed verification, meaning that it could not be located in the source translation. & Remove uncertain alignments. \\
Very short reference & Exclude if any English reference passage was under 100 characters. & Avoid fragments too short for meaningful comparison. \\
Near-identical references & Exclude if any pair of references had at least 90\% normalized character 3-gram Jaccard similarity. & Avoid formulaic or duplicated translations that collapse the reference envelope. \\
Length imbalance & Exclude if the longest reference was more than twice the length of the shortest. & Remove likely partial alignments or severe extraction mismatches. \\
Null-character corruption & Exclude if any text field contained null-character corruption. & Remove encoding or extraction artifacts. \\
Internal duplication & Exclude if internal segment duplication suggested extraction error. & Remove malformed assembled passages. \\
Near-duplicate Pali & Remove later passages when Pali 3-gram Jaccard similarity exceeded 85\%, retaining the first passage by alphabetical passage identifier. & Reduce formulaic repetition and overrepresentation. \\
\end{xltabular}
\endgroup

The first seven passage-level filters were non-exclusive and reduced 4,003 candidates to 1,874 passages. The final near-duplicate removal pass removed 174 additional passages and produced the final 1,700-passage test set.

The final test set contains 1,700 Pali passages: 738 SN passages, 553 MN passages, and 409 AN passages, spanning 8,389 source segments. The Pali source side contains 645,876 characters, 70,846 whitespace-delimited words, and approximately 345,000 model tokens using the \texttt{cl100k\_\allowbreak{}base} encoding. As a rough scale comparison, this is on the order of 200-250 printed A4 pages at 12-point type with 1.5 line spacing, depending on margins and font. Median source length is 34 whitespace-delimited Pali words, with a mean of 41.7. Each item also contains the corresponding English reference translations by Bhikkhu Sujato, Bhikkhu Thanissaro, and Bhikkhu Bodhi [SuttaCentral n.d.]; [Thanissaro Bhikkhu n.d.]; [Bodhi 2000]; [Bodhi 2012]; [Ñāṇamoli and Bodhi 1995].

The three-reference design is central to the method. Because Pali Buddhist texts admit legitimate variation in register, terminology, compression, and interpretation, no single reference translation is treated as definitive. Instead, the spread among the three human translations defines a local reference envelope against which candidate translations can be compared. SuttaCentral also functions as digital humanities infrastructure: its structured texts, aligned translations, segment identifiers, and public reading/search environment make passage-level computational comparison possible.

\subsection{Translator Models}

Selected LLMs were tasked with translating the Pali source dataset into English. The main translator panel contains four flagship general-purpose models from distinct provider families:

\par\medskip\noindent\textbf{Table 2. Primary translator model panel.}\par\smallskip
\begingroup
\small
\setlength{\tabcolsep}{3pt}
\begin{xltabular}{\textwidth}{@{}>{\RaggedRight\arraybackslash}p{0.24\textwidth}Y@{}}
\toprule
\textbf{Provider} & \textbf{Model ID (OpenRouter)} \\
\midrule
\endfirsthead
\toprule
\textbf{Provider} & \textbf{Model ID (OpenRouter)} \\
\midrule
\endhead
\midrule
\multicolumn{2}{r}{\footnotesize continued on next page}\\
\endfoot
\bottomrule
\endlastfoot
OpenAI & \texttt{openai/\allowbreak{}gpt-\allowbreak{}5.\allowbreak{}5} \\
Anthropic & \texttt{anthropic/\allowbreak{}claude-\allowbreak{}sonnet-\allowbreak{}4.\allowbreak{}6} \\
Google & \texttt{google/\allowbreak{}gemini-\allowbreak{}3.\allowbreak{}1-\allowbreak{}pro-\allowbreak{}preview} \\
xAI & \texttt{x-\allowbreak{}ai/\allowbreak{}grok-\allowbreak{}4.\allowbreak{}3} \\
\end{xltabular}
\endgroup

All models were prompted with the same zero-shot translation prompt and asked to return JavaScript Object Notation (JSON) mapping passage IDs to English translations. No examples, reference translations, or wider discourse context were provided. The full translation prompt is reproduced in Appendix A.

Translation calls were made through the OpenRouter chat-completions application programming interface (API) using temperature 0 and JSON response mode. The translation script batched passages by a target budget of approximately 3,000 source-side tokens per request. To keep results comparable, reasoning was explicitly requested as \texttt{none} where supported; for Gemini 3.1 Pro, the request was mapped to minimal reasoning because the provider mapping did not support a literal \texttt{none} setting. This controls the requested provider-side reasoning configuration and token accounting, but should not be interpreted as proof that a model performed no internal reasoning at all. GPT-5.5 translated on May 4, 2026; Claude Sonnet 4.6, Gemini 3.1 Pro, and Grok 4.3 translated on May 8, 2026.

\subsection{Embedding Drift and Outlier Triage}

Each model produces an English candidate translation, and this English output is compared with the three English human reference translations. The candidate translation and each reference translation are embedded with Qwen3 Embedding 8B [Zhang et al. 2025], selected because its technical report reports strong performance on broad text-embedding benchmarks, including the Massive Text Embedding Benchmark (MTEB) [Muennighoff et al. 2023]; [Zhang et al. 2025]. Pali source text is not embedded for the drift calculation. For each passage, we compute similarity from the candidate to each reference and to the reference centroid.

The main triage statistic is normalized drift, calculated in three steps. First, the three English reference translations for a passage are embedded and averaged to form a reference centroid. Second, the Euclidean distance between the candidate translation embedding and this centroid is computed. Third, that candidate-to-centroid distance is divided by the average Euclidean distance between each human reference embedding and the same centroid. Thus, a normalized drift value of \texttt{1.\allowbreak{}0} means that the candidate is about as far from the center of the reference set as the references are from one another on average; values above \texttt{1.\allowbreak{}0} indicate increasing distance outside the local reference envelope.

Normalized drift is interpreted as a triage score, not an error label. A high-drift translation is outside the local human-reference envelope and therefore deserves adjudication, but it may still be a valid translation variation. Conversely, a low-drift translation is not guaranteed to be error-free, though (as the validation results below indicate) the risk of a major error is substantially lower in the low-drift range. Consequently, passages where normalized drift exceeded 1.5 were selected for LLM adjudication. Results are stratified into five drift bands: \texttt{0.\allowbreak{}0-\allowbreak{}1.\allowbreak{}0}, \texttt{1.\allowbreak{}0-\allowbreak{}1.\allowbreak{}5}, \texttt{1.\allowbreak{}5-\allowbreak{}2.\allowbreak{}0}, \texttt{2.\allowbreak{}0-\allowbreak{}3.\allowbreak{}0}, and \texttt{\textgreater{}3.\allowbreak{}0}.

\subsection{LLM Judge Panel}

The primary LLM judge panel is independent of the translator panel:

\par\medskip\noindent\textbf{Table 3. Primary LLM judge panel.}\par\smallskip
\begingroup
\small
\setlength{\tabcolsep}{3pt}
\begin{xltabular}{\textwidth}{@{}>{\RaggedRight\arraybackslash}p{0.24\textwidth}Y@{}}
\toprule
\textbf{Judge} & \textbf{Model ID} \\
\midrule
\endfirsthead
\toprule
\textbf{Judge} & \textbf{Model ID} \\
\midrule
\endhead
\midrule
\multicolumn{2}{r}{\footnotesize continued on next page}\\
\endfoot
\bottomrule
\endlastfoot
DeepSeek & \texttt{deepseek/\allowbreak{}deepseek-\allowbreak{}v4-\allowbreak{}pro} \\
Kimi & \texttt{moonshotai/\allowbreak{}kimi-\allowbreak{}k2.\allowbreak{}6} \\
Qwen & \texttt{qwen/\allowbreak{}qwen3.\allowbreak{}6-\allowbreak{}max-\allowbreak{}preview} \\
\end{xltabular}
\endgroup

Judges are blinded to the translator model identity. Each judge sees the Pali source, the three reference translations and the candidate translation. Each translation is judged independently, one passage per API call. Judge calls were also made through OpenRouter, with the same API parameters described above. The judge prompt requested a fixed JSON object including the top-level label (e.g. VALID\_VARIATION, MINOR\_ERROR), error category, severity and confidence. Judge outputs were parsed as JSON and validated against the allowed label, category, severity, and confidence sets. Invalid outputs were retried up to five times with exponential backoff.

\subsection{Error Labels}

The LLM judges use five top-level labels as described in Table 4.

\par\medskip\noindent\textbf{Table 4. Top-level adjudication labels and operational definitions.}\par\smallskip
\begingroup
\small
\setlength{\tabcolsep}{3pt}
\begin{xltabular}{\textwidth}{@{}>{\RaggedRight\arraybackslash}p{0.24\textwidth}Y@{}}
\toprule
\textbf{Label} & \textbf{Definition} \\
\midrule
\endfirsthead
\toprule
\textbf{Label} & \textbf{Definition} \\
\midrule
\endhead
\midrule
\multicolumn{2}{r}{\footnotesize continued on next page}\\
\endfoot
\bottomrule
\endlastfoot
\texttt{VALID\_\allowbreak{}VARIATION} & A defensible translation; differences mainly reflect style, register, terminology, syntax, compression, expansion, or interpretive variation. \\
\texttt{MINOR\_\allowbreak{}ERROR} & A real, correction-worthy translation problem that does not substantially change the passage's main meaning. The core agents, actions, negation, doctrinal point, logical relation, and practical implication remain recoverable. \\
\texttt{MAJOR\_\allowbreak{}ERROR} & A translation problem that materially changes, obscures, or loses the passage's meaning. A reader relying only on the candidate translation would be materially misled. \\
\texttt{REFERENCE\_\allowbreak{}OR\_\allowbreak{}ALIGNMENT\_\allowbreak{}PROBLEM} & The references or passage alignment appear to be the main source of discrepancy. \\
\texttt{UNSURE\_\allowbreak{}CONTEXT\_\allowbreak{}NEEDED} & The item cannot be fairly adjudicated without wider sutta context. \\
\end{xltabular}
\endgroup

The operational boundary between minor and major errors is reader consequence. If a careful translator would correct the candidate but a reader would still recover the main meaning, the item is a \texttt{MINOR\_\allowbreak{}ERROR}. If a reader would misunderstand who did what, whether something was affirmed or denied, what was included in a list, what was the correct number, how clauses relate, or what doctrinal point is being made, the item is a \texttt{MAJOR\_\allowbreak{}ERROR}.

Error categories are assigned for minor and major errors:

\par\medskip\noindent\textbf{Table 5. Error taxonomy used for minor and major translation errors.}\par\smallskip
\begingroup
\small
\setlength{\tabcolsep}{3pt}
\begin{xltabular}{\textwidth}{@{}Y@{}}
\toprule
\textbf{Error category} \\
\midrule
\endfirsthead
\toprule
\textbf{Error category} \\
\midrule
\endhead
\midrule
\multicolumn{1}{r}{\footnotesize continued on next page}\\
\endfoot
\bottomrule
\endlastfoot
\texttt{OMISSION\_\allowbreak{}OR\_\allowbreak{}TRUNCATION} \\
\texttt{ADDITION\_\allowbreak{}OR\_\allowbreak{}HALLUCINATION} \\
\texttt{NEGATION\_\allowbreak{}OR\_\allowbreak{}POLARITY\_\allowbreak{}ERROR} \\
\texttt{AGENT\_\allowbreak{}OR\_\allowbreak{}ROLE\_\allowbreak{}ERROR} \\
\texttt{DOCTRINAL\_\allowbreak{}TERM\_\allowbreak{}ERROR} \\
\texttt{LIST\_\allowbreak{}OR\_\allowbreak{}NUMBER\_\allowbreak{}ERROR} \\
\texttt{UNRESOLVED\_\allowbreak{}SOURCE\_\allowbreak{}ELLIPSIS} \\
\texttt{WRONG\_\allowbreak{}RELATION\_\allowbreak{}OR\_\allowbreak{}LOGIC} \\
\texttt{OVERLY\_\allowbreak{}FREE\_\allowbreak{}PARAPHRASE} \\
\texttt{GRAMMAR\_\allowbreak{}OR\_\allowbreak{}FLUENCY\_\allowbreak{}PROBLEM} \\
\texttt{OTHER} \\
\end{xltabular}
\endgroup

For the three-judge panel, the majority aggregation rule is summarized in Appendix Table E1: two major-error labels produce a panel major error, two error labels without a major-error majority produce a panel minor error, and two valid-variation labels produce a panel valid variation.

\subsection{Author-Adjudicated Validation Set}

To calibrate the relation between embedding drift and translation error, and to measure how well the LLM judges reproduce a human adjudicator's labels, we created a 300-item author-adjudicated validation set where both the LLM judge panel and one of the authors adjudicated passages.

This validation sample was stratified by normalized drift so that it included both low-drift translations, where major errors were expected to be uncommon, and high-drift translations, where the audit method expected errors to be enriched. The sample contains 100 items from the \texttt{0.\allowbreak{}0-\allowbreak{}1.\allowbreak{}0} band, 100 from \texttt{1.\allowbreak{}0-\allowbreak{}1.\allowbreak{}5}, 50 from \texttt{1.\allowbreak{}5-\allowbreak{}2.\allowbreak{}0}, and 50 from \texttt{\textgreater{}2.\allowbreak{}0}. The author adjudicator has formal training in Pali, but this validation subset is not described as independent expert validation.

\subsection{Main LLM Adjudication Subset}

For the primary model-level analysis, all translation instances with \texttt{normalized\_\allowbreak{}drift \textgreater{} 1.\allowbreak{}5} were selected and sent to the three-judge panel. This subset contains 1,203 translation instances: 224 from GPT-5.5, 302 from Claude Sonnet 4.6, 257 from Gemini 3.1 Pro, and 420 from Grok 4.3.

The result-set analysis uses the majority-panel label as the primary adjudicated outcome. Error-category analysis is restricted to instances labeled \texttt{MINOR\_\allowbreak{}ERROR} or \texttt{MAJOR\_\allowbreak{}ERROR} by the majority panel. If judges agreed that an item was an error but assigned different error categories, we used the category chosen by the largest number of error-labeling judges. Error rates reported for the LLM-adjudicated result set refer only to this high-drift subset (\texttt{normalized\_\allowbreak{}drift \textgreater{} 1.\allowbreak{}5}). The author-validation sample is used separately to interpret the likely error burden below the full-adjudication threshold. Where confidence intervals are reported, they are two-sided 95\% Wilson score intervals for binomial proportions [Wilson 1927].

\subsection{Automatic metrics}

For descriptive comparison across translator models, we also report automatic metrics over the full 1,700-passage test set. \texttt{sim\_\allowbreak{}best} is the mean highest embedding similarity between a candidate translation and any of the three English references. COMET average is the mean COMET score against the three references, included as a learned MT metric from the COMET and WMT metrics tradition [Rei et al. 2020]; [Rei et al. 2022]; [Freitag et al. 2022]. Length ratio is the candidate translation length divided by the average reference length for the same passage, averaged across passages. These metrics are used as descriptive context and outlier-triage support, not as substitutes for adjudicated error labels.

\section{Results}

\subsection{Automatic Translation Metrics}

All four primary translators completed the full 1,700-passage test set. Gemini 3.1 Pro had the highest mean semantic similarity to the nearest reference, while Grok 4.3 had the largest normalized drift and highest outlier rate.

\par\medskip\noindent\textbf{Table 6. Automatic translation metrics and outlier counts for the four primary translators.}\par\smallskip
\begingroup
\scriptsize
\setlength{\tabcolsep}{2pt}
\begin{xltabular}{\textwidth}{@{}Y Y Y Y Y Y Y@{}}
\toprule
\textbf{Translator} & \textbf{\texttt{sim\_\allowbreak{}best}} & \textbf{Mean normalized drift} & \textbf{Length ratio} & \textbf{COMET avg} & \textbf{\texttt{\textgreater{}1.\allowbreak{}5} outliers} & \textbf{\texttt{\textgreater{}2.\allowbreak{}0} outliers} \\
\midrule
\endfirsthead
\toprule
\textbf{Translator} & \textbf{\texttt{sim\_\allowbreak{}best}} & \textbf{Mean normalized drift} & \textbf{Length ratio} & \textbf{COMET avg} & \textbf{\texttt{\textgreater{}1.\allowbreak{}5} outliers} & \textbf{\texttt{\textgreater{}2.\allowbreak{}0} outliers} \\
\midrule
\endhead
\midrule
\multicolumn{7}{r}{\footnotesize continued on next page}\\
\endfoot
\bottomrule
\endlastfoot
GPT-5.5 & 0.937 & 1.107 & 1.088 & 0.719 & 224 & 84 \\
Claude Sonnet 4.6 & 0.936 & 1.172 & 1.102 & 0.716 & 302 & 112 \\
Gemini 3.1 Pro & 0.947 & 1.151 & 0.942 & 0.728 & 257 & 80 \\
Grok 4.3 & 0.928 & 1.389 & 1.056 & 0.700 & 420 & 201 \\
\end{xltabular}
\endgroup

The \texttt{\textgreater{}1.\allowbreak{}5} result set contains 1,203 translation instances. Appendix Table E2 gives the simple result-set counts by translator; drift-band counts are reported in the main adjudication table below.

\subsection{Author Validation Results}

The author-adjudicated validation set (n = 300) contains 227 valid variations, 58 minor errors, and 15 major errors; label counts are reported in Appendix Table E3.

Error rates by drift band show that minor errors occur across all bands, while major errors concentrate strongly above \texttt{2.\allowbreak{}0}.

\par\medskip\noindent\textbf{Table 7. Author-adjudicated validation outcomes by drift band.}\par\smallskip
\begingroup
\scriptsize
\setlength{\tabcolsep}{2pt}
\begin{xltabular}{\textwidth}{@{}Y Y Y Y Y Y Y@{}}
\toprule
\textbf{Drift band} & \textbf{n} & \textbf{Valid} & \textbf{Minor error} & \textbf{Major error} & \textbf{Major 95\% confidence interval} & \textbf{Any error} \\
\midrule
\endfirsthead
\toprule
\textbf{Drift band} & \textbf{n} & \textbf{Valid} & \textbf{Minor error} & \textbf{Major error} & \textbf{Major 95\% confidence interval} & \textbf{Any error} \\
\midrule
\endhead
\midrule
\multicolumn{7}{r}{\footnotesize continued on next page}\\
\endfoot
\bottomrule
\endlastfoot
\texttt{0.\allowbreak{}0-\allowbreak{}1.\allowbreak{}0} & 100 & 85.0\% & 14.0\% & 1.0\% & 0.2-5.4\% & 15.0\% \\
\texttt{1.\allowbreak{}0-\allowbreak{}1.\allowbreak{}5} & 100 & 77.0\% & 21.0\% & 2.0\% & 0.6-7.0\% & 23.0\% \\
\texttt{1.\allowbreak{}5-\allowbreak{}2.\allowbreak{}0} & 50 & 74.0\% & 24.0\% & 2.0\% & 0.4-10.5\% & 26.0\% \\
\texttt{\textgreater{}2.\allowbreak{}0} & 50 & 56.0\% & 22.0\% & 22.0\% & 12.8-35.2\% & 44.0\% \\
\end{xltabular}
\endgroup

In the author-validation sample, higher normalized drift is not a clean boundary for all errors: minor errors are present in every sampled band. Major errors are uncommon below \texttt{2.\allowbreak{}0} in this sample and more frequent above \texttt{2.\allowbreak{}0}.

Below the main \texttt{1.\allowbreak{}5} adjudication threshold, the author-validation set contains 3 major errors in 200 instances, or 1.5\% with a Wilson 95\% interval of 0.5-4.3\%. This interval is wide enough that below-threshold major errors cannot be ruled out, but they are uncommon in the validation sample.

Exploratory per-translator counts below the \texttt{1.\allowbreak{}5} threshold are small, with 50 validation instances per translator. GPT-5.5 had 0 major and 8 minor errors, Claude Sonnet 4.6 had 1 major and 5 minor errors, Gemini 3.1 Pro had 1 major and 7 minor errors, and Grok 4.3 had 1 major and 15 minor errors. These counts suggest that Grok's weaker performance is not confined to the high-drift tail, but the cell sizes are too small for strong per-model claims below threshold.

\subsection{LLM Judge Calibration on the Validation Set}

The primary three-judge panel was evaluated against the author labels on the 300 validation instances.

\par\medskip\noindent\textbf{Table 8. LLM judge calibration against author-validation labels.}\par\smallskip
\begingroup
\fontsize{6.8}{7.8}\selectfont
\setlength{\tabcolsep}{1.5pt}
\begin{xltabular}{\textwidth}{@{}Y Y Y Y Y Y Y Y@{}}
\toprule
\textbf{Judge} & \textbf{Exact agreement} & \textbf{Major precision} & \textbf{Major recall} & \textbf{Major F1} & \textbf{Any-error precision} & \textbf{Any-error recall} & \textbf{Any-error F1} \\
\midrule
\endfirsthead
\toprule
\textbf{Judge} & \textbf{Exact agreement} & \textbf{Major precision} & \textbf{Major recall} & \textbf{Major F1} & \textbf{Any-error precision} & \textbf{Any-error recall} & \textbf{Any-error F1} \\
\midrule
\endhead
\midrule
\multicolumn{8}{r}{\footnotesize continued on next page}\\
\endfoot
\bottomrule
\endlastfoot
DeepSeek V4 Pro & 77.7\% & 44.4\% & 53.3\% & 48.5\% & 62.7\% & 64.4\% & 63.5\% \\
Kimi K2.6 & 77.0\% & 56.2\% & 60.0\% & 58.1\% & 66.7\% & 32.9\% & 44.0\% \\
Qwen3.6 Max Preview & 82.7\% & 64.7\% & 73.3\% & 68.8\% & 82.6\% & 52.1\% & 63.9\% \\
\textbf{Majority panel} & \textbf{80.0\%} & \textbf{66.7\%} & \textbf{66.7\%} & \textbf{66.7\%} & \textbf{72.5\%} & \textbf{50.7\%} & \textbf{59.7\%} \\
\end{xltabular}
\endgroup

The panel is better calibrated for major errors than for minor errors. It detects two-thirds of author-labeled major errors with two-thirds precision, but it misses many minor errors by classifying them as valid variation.

The majority-panel confusion matrix is reported in Appendix Table E4. Although the adjudication schema allowed \texttt{UNSURE\_\allowbreak{}CONTEXT\_\allowbreak{}NEEDED} and \texttt{REFERENCE\_\allowbreak{}OR\_\allowbreak{}ALIGNMENT\_\allowbreak{}PROBLEM}, these labels had zero author-label counts and zero majority-panel counts in the validation set, so the confusion matrix reduces to the three substantive translation-quality labels.

No author-labeled valid variation was upgraded to major error by the majority panel, and no author-labeled major error was classified as valid variation by the majority panel. Most disagreement therefore concerns the boundary between valid variation and minor error, and between minor and major severity.

Cohen's kappa [Cohen 1960] shows the same pattern in chance-corrected terms. Agreement between the author labels and majority-panel labels is moderate for the three-class label task (\texttt{kappa = 0.\allowbreak{}422}) and for the valid-versus-any-error binary task (\texttt{kappa = 0.\allowbreak{}496}), but stronger for the major-error-versus-not-major binary task (\texttt{kappa = 0.\allowbreak{}649}).

Pairwise exact agreement among the three primary judges on the 300-item validation set was 81.3\% for DeepSeek/Kimi, 85.3\% for DeepSeek/Qwen, and 86.3\% for Kimi/Qwen. The corresponding three-class Cohen's kappa values are 0.424 for DeepSeek/Kimi, 0.573 for DeepSeek/Qwen, and 0.443 for Kimi/Qwen.

On the full high-drift result set, pairwise judge agreement was lower: 75.6\% for DeepSeek/Kimi, 74.7\% for DeepSeek/Qwen, and 81.5\% for Kimi/Qwen. This is consistent with the result set being enriched for more difficult or ambiguous cases.

\subsection{Full Result-Set Adjudication by Drift Band}

The three-judge panel adjudicated all 1,203 translation instances with \texttt{normalized\_\allowbreak{}drift \textgreater{} 1.\allowbreak{}5}. Across this high-drift result set, 865 instances were classified as valid variations, 169 as minor errors, and 169 as major errors.

Error frequency increased monotonically with drift band. The increase was driven mainly by major errors: only 7.9\% of \texttt{1.\allowbreak{}5-\allowbreak{}2.\allowbreak{}0} instances were labeled major errors, compared with 13.6\% of \texttt{2.\allowbreak{}0-\allowbreak{}3.\allowbreak{}0} instances and 51.6\% of \texttt{\textgreater{}3.\allowbreak{}0} instances.

\par\medskip\noindent\textbf{Table 9. Majority-panel adjudication outcomes for all high-drift instances by drift band.}\par\smallskip
\begingroup
\scriptsize
\setlength{\tabcolsep}{2pt}
\begin{xltabular}{\textwidth}{@{}Y Y Y Y Y Y Y@{}}
\toprule
\textbf{Drift band} & \textbf{n} & \textbf{Valid variation} & \textbf{Minor error} & \textbf{Major error} & \textbf{Major 95\% confidence interval} & \textbf{Any error} \\
\midrule
\endfirsthead
\toprule
\textbf{Drift band} & \textbf{n} & \textbf{Valid variation} & \textbf{Minor error} & \textbf{Major error} & \textbf{Major 95\% confidence interval} & \textbf{Any error} \\
\midrule
\endhead
\midrule
\multicolumn{7}{r}{\footnotesize continued on next page}\\
\endfoot
\bottomrule
\endlastfoot
\texttt{1.\allowbreak{}5-\allowbreak{}2.\allowbreak{}0} & 726 & 79.9\% & 12.3\% & 7.9\% & 6.1-10.0\% & 20.1\% \\
\texttt{2.\allowbreak{}0-\allowbreak{}3.\allowbreak{}0} & 353 & 68.6\% & 17.8\% & 13.6\% & 10.4-17.6\% & 31.4\% \\
\texttt{\textgreater{}3.\allowbreak{}0} & 124 & 34.7\% & 13.7\% & 51.6\% & 42.9-60.2\% & 65.3\% \\
\end{xltabular}
\endgroup

Most translations in the \texttt{1.\allowbreak{}5-\allowbreak{}2.\allowbreak{}0} band remain valid, while the share of major errors rises sharply in the extreme tail. In the \texttt{\textgreater{}3.\allowbreak{}0} band, a translation is more likely to contain an error than to be labeled a valid variation.

\subsection{Full Result-Set Adjudication by Translator}

The four translator models differed substantially in the number of high-drift instances and in the adjudicated error composition of those instances.

\par\medskip\noindent\textbf{Table 10. Majority-panel adjudication outcomes for high-drift instances by translator.}\par\smallskip
\begingroup
\scriptsize
\setlength{\tabcolsep}{2pt}
\begin{xltabular}{\textwidth}{@{}Y Y Y Y Y Y Y@{}}
\toprule
\textbf{Translator} & \textbf{High-drift n} & \textbf{Valid variation} & \textbf{Minor error} & \textbf{Major error} & \textbf{Major 95\% confidence interval} & \textbf{Any error} \\
\midrule
\endfirsthead
\toprule
\textbf{Translator} & \textbf{High-drift n} & \textbf{Valid variation} & \textbf{Minor error} & \textbf{Major error} & \textbf{Major 95\% confidence interval} & \textbf{Any error} \\
\midrule
\endhead
\midrule
\multicolumn{7}{r}{\footnotesize continued on next page}\\
\endfoot
\bottomrule
\endlastfoot
GPT-5.5 & 224 & 87.9\% & 8.0\% & 4.0\% & 2.1-7.5\% & 12.1\% \\
Claude Sonnet 4.6 & 302 & 77.5\% & 13.9\% & 8.6\% & 5.9-12.3\% & 22.5\% \\
Gemini 3.1 Pro & 257 & 79.0\% & 14.0\% & 7.0\% & 4.5-10.8\% & 21.0\% \\
Grok 4.3 & 420 & 55.0\% & 17.4\% & 27.6\% & 23.6-32.1\% & 45.0\% \\
\end{xltabular}
\endgroup

GPT-5.5 had the lowest point estimate for high-drift major-error burden in this result set, but the Wilson intervals overlap among GPT-5.5, Gemini 3.1 Pro, and Claude Sonnet 4.6. The clearest model-level separation is between Grok 4.3 and the other three translators: Grok 4.3 had both the largest number of high-drift instances and a much higher major-error rate within those instances.

These are not total corpus error rates, because translations below the \texttt{1.\allowbreak{}5} drift threshold were not fully adjudicated. They are direct counts of adjudicated errors in the high-drift tail. The author-validation sample indicates that minor errors also occur below \texttt{1.\allowbreak{}5}, while major errors below \texttt{1.\allowbreak{}5} were uncommon in the validation sample.

Appendix Table E5 expresses these same observed high-drift error counts as percentages of the full 1,700-passage dataset. Because minor errors also occur below the adjudication threshold, and major errors cannot be ruled out below it, these percentages should be interpreted as lower-bound estimates.

Also, it should be taken into account that the panel aggregation rule assigns \texttt{MINOR\_\allowbreak{}ERROR} when all three judges disagree as \texttt{VALID\_\allowbreak{}VARIATION}, \texttt{MINOR\_\allowbreak{}ERROR}, and \texttt{MAJOR\_\allowbreak{}ERROR}. This occurred in 30 of 1,203 result-set instances, or 2.5\% of the high-drift set and 17.8\% of panel-minor labels.

\subsection{Model-by-Band Structure}

The drift/error relationship is not identical across translators. Grok 4.3 accounts for most of the extreme-tail risk: 82 of its translations fall above \texttt{3.\allowbreak{}0}, and 74.4\% of those are major errors. GPT-5.5 has 20 translations above \texttt{3.\allowbreak{}0}, but none were labeled major errors by the panel; most were valid variations or minor issues. Claude and Gemini have comparatively few \texttt{\textgreater{}3.\allowbreak{}0} cases.

\par\medskip\noindent\textbf{Table 11. Majority-panel adjudication outcomes by translator and drift band.}\par\smallskip
\begingroup
\scriptsize
\setlength{\tabcolsep}{2pt}
\begin{xltabular}{\textwidth}{@{}Y Y Y Y Y Y@{}}
\toprule
\textbf{Translator} & \textbf{Band} & \textbf{n} & \textbf{Valid variation} & \textbf{Minor error} & \textbf{Major error} \\
\midrule
\endfirsthead
\toprule
\textbf{Translator} & \textbf{Band} & \textbf{n} & \textbf{Valid variation} & \textbf{Minor error} & \textbf{Major error} \\
\midrule
\endhead
\midrule
\multicolumn{6}{r}{\footnotesize continued on next page}\\
\endfoot
\bottomrule
\endlastfoot
GPT-5.5 & \texttt{1.\allowbreak{}5-\allowbreak{}2.\allowbreak{}0} & 140 & 92.9\% & 5.0\% & 2.1\% \\
GPT-5.5 & \texttt{2.\allowbreak{}0-\allowbreak{}3.\allowbreak{}0} & 64 & 84.4\% & 6.2\% & 9.4\% \\
GPT-5.5 & \texttt{\textgreater{}3.\allowbreak{}0} & 20 & 65.0\% & 35.0\% & 0.0\% \\
Claude Sonnet 4.6 & \texttt{1.\allowbreak{}5-\allowbreak{}2.\allowbreak{}0} & 190 & 83.2\% & 9.5\% & 7.4\% \\
Claude Sonnet 4.6 & \texttt{2.\allowbreak{}0-\allowbreak{}3.\allowbreak{}0} & 94 & 66.0\% & 23.4\% & 10.6\% \\
Claude Sonnet 4.6 & \texttt{\textgreater{}3.\allowbreak{}0} & 18 & 77.8\% & 11.1\% & 11.1\% \\
Gemini 3.1 Pro & \texttt{1.\allowbreak{}5-\allowbreak{}2.\allowbreak{}0} & 177 & 80.2\% & 13.0\% & 6.8\% \\
Gemini 3.1 Pro & \texttt{2.\allowbreak{}0-\allowbreak{}3.\allowbreak{}0} & 76 & 77.6\% & 15.8\% & 6.6\% \\
Gemini 3.1 Pro & \texttt{\textgreater{}3.\allowbreak{}0} & 4 & 50.0\% & 25.0\% & 25.0\% \\
Grok 4.3 & \texttt{1.\allowbreak{}5-\allowbreak{}2.\allowbreak{}0} & 219 & 68.5\% & 18.7\% & 12.8\% \\
Grok 4.3 & \texttt{2.\allowbreak{}0-\allowbreak{}3.\allowbreak{}0} & 119 & 56.3\% & 21.0\% & 22.7\% \\
Grok 4.3 & \texttt{\textgreater{}3.\allowbreak{}0} & 82 & 17.1\% & 8.5\% & 74.4\% \\
\end{xltabular}
\endgroup

Among high-drift instances, automatic metrics correlate with panel-major labels in the expected directions. The strongest observed point-biserial correlations with major-error status are COMET average (\texttt{r = -\allowbreak{}0.\allowbreak{}527}), COMET best (\texttt{r = -\allowbreak{}0.\allowbreak{}526}), normalized drift (\texttt{r = 0.\allowbreak{}456}), chrF (\texttt{r = -\allowbreak{}0.\allowbreak{}445}), centroid similarity (\texttt{r = -\allowbreak{}0.\allowbreak{}438}), nearest-reference similarity (\texttt{r = -\allowbreak{}0.\allowbreak{}428}), and BLEU (\texttt{r = -\allowbreak{}0.\allowbreak{}347}). Length ratio is much weaker (\texttt{r = -\allowbreak{}0.\allowbreak{}120}), and human-reference variance is near zero (\texttt{r = 0.\allowbreak{}029}). These correlations are descriptive and restricted to the already high-drift result set.

\subsection{Error Categories}

The most common major-error category was omission or truncation, followed by doctrinal term errors, agent or role errors, additions or hallucinations, and negation or polarity errors.

\par\medskip\noindent\textbf{Table 12. Majority-panel major-error categories in the high-drift result set.}\par\smallskip
\begingroup
\small
\setlength{\tabcolsep}{3pt}
\begin{xltabular}{\textwidth}{@{}>{\RaggedRight\arraybackslash}p{0.24\textwidth}Y@{}}
\toprule
\textbf{Major-error category} & \textbf{Count} \\
\midrule
\endfirsthead
\toprule
\textbf{Major-error category} & \textbf{Count} \\
\midrule
\endhead
\midrule
\multicolumn{2}{r}{\footnotesize continued on next page}\\
\endfoot
\bottomrule
\endlastfoot
\texttt{OMISSION\_\allowbreak{}OR\_\allowbreak{}TRUNCATION} & 52 \\
\texttt{DOCTRINAL\_\allowbreak{}TERM\_\allowbreak{}ERROR} & 35 \\
\texttt{AGENT\_\allowbreak{}OR\_\allowbreak{}ROLE\_\allowbreak{}ERROR} & 22 \\
\texttt{ADDITION\_\allowbreak{}OR\_\allowbreak{}HALLUCINATION} & 20 \\
\texttt{NEGATION\_\allowbreak{}OR\_\allowbreak{}POLARITY\_\allowbreak{}ERROR} & 18 \\
\texttt{UNRESOLVED\_\allowbreak{}SOURCE\_\allowbreak{}ELLIPSIS} & 11 \\
\texttt{LIST\_\allowbreak{}OR\_\allowbreak{}NUMBER\_\allowbreak{}ERROR} & 5 \\
\texttt{WRONG\_\allowbreak{}RELATION\_\allowbreak{}OR\_\allowbreak{}LOGIC} & 4 \\
\texttt{GRAMMAR\_\allowbreak{}OR\_\allowbreak{}FLUENCY\_\allowbreak{}PROBLEM} & 1 \\
\texttt{OTHER} & 1 \\
\end{xltabular}
\endgroup

Minor errors had a different profile. Unresolved source ellipsis and doctrinal terminology dominated minor-error labels, while omission and polarity errors were less common as minor errors than as major errors. In this context, unresolved ellipsis refers to cases where repetitive canonical material is abbreviated with ellipsis marks or equivalent omission conventions. Such abbreviation is common in Pali canonical texts because formulas, lists, and repeated doctrinal sequences recur across many passages; a translator may need to infer the omitted repeated material from context. The minor-error category counts are reported in Appendix Table E6.

This distinction is substantively important. The common minor-error profile reflects translation awkwardness, unresolved repetition, and terminology choices that are correction-worthy but often recoverable. The major-error profile reflects failures that alter the content available to the reader: missing material, reversed or distorted polarity, wrong agents, added content, or doctrinally misleading renderings.

The ellipsis policy has a visible effect on minor-error rates. \texttt{UNRESOLVED\_\allowbreak{}SOURCE\_\allowbreak{}ELLIPSIS} accounts for 60 of 169 minor errors and 11 of 169 major errors. If unresolved-ellipsis cases are excluded from the error count, the high-drift result set changes from 14.0\% minor, 14.0\% major, and 28.1\% any error to 9.1\% minor, 13.1\% major, and 22.2\% any error. Thus the main major-error findings are not driven by ellipsis policy, but the minor-error rate is materially affected by it.

Major-error categories also differ by translator. Grok 4.3's surplus major errors are dominated by omission or truncation (48 cases), followed by doctrinal term errors (19), negation or polarity errors (13), additions or hallucinations (13), agent or role errors (10), and unresolved ellipsis (10). Claude Sonnet 4.6's largest major-error category is doctrinal terminology (10 cases). Gemini 3.1 Pro's major errors are led by agent or role errors (6 cases), while GPT-5.5 has only nine high-drift major errors spread across several categories.

\subsection{Shared and Model-Specific Failures}

Within the high-drift judged set, major errors were mostly model-specific rather than shared by all translators on the same Pali passage. Among high-drift judged items, 150 unique Pali passages had at least one model with a major error. Of these, 132 passages had a major error in only one judged translator, 17 in two judged translators, and 1 in three judged translators. No passage had a high-drift judged major error in all four translator models.

For any error, including minor errors, overlap was broader but still far from universal. There were 258 unique Pali passages with at least one high-drift model error: 195 involved one translator, 47 involved two translators, 15 involved three translators, and 1 involved all four translators. Appendix Table E7 reports these any-error overlap counts.

This pattern suggests that many failures are not simply properties of intrinsically impossible source passages. However, the claim is conditional on the high-drift adjudication set. For the 150 passages with at least one high-drift major error, 278 of the 600 possible passage-model pairs were judged, while 322 model translations were below the \texttt{1.\allowbreak{}5} threshold and not judged. A stronger model-specificity claim would require adjudicating those below-threshold companion translations.

Appendix Table E8 gives the clearest shared-major-error examples, including the only three-major passage (\texttt{sn4.\allowbreak{}24:\allowbreak{}10}) and the only two-major passage where both remaining translators were also judged erroneous, though only at minor severity (\texttt{an4.\allowbreak{}24:\allowbreak{}5}). In that appendix table, "below threshold / not judged" means that the model's translation for the passage had \texttt{normalized\_\allowbreak{}drift \textless{}= 1.\allowbreak{}5}, did not enter the high-drift result set, and therefore has no LLM-judge label; it should not be read as valid.

The following validation-set example shows why numerical outlier detection needs philological adjudication. In \texttt{sn4.\allowbreak{}25:\allowbreak{}2}, Māra's daughters address Māra after his failed attempt to sway the Buddha. They ask which man he is grieving over and promise to bind and bring that man, the Buddha, under Māra's power. The candidate instead makes Māra himself the object of binding and control.

\par\medskip\noindent\textbf{Table 13. Worked example of an author- and panel-confirmed major agent/role error (\texttt{normalized\_\allowbreak{}drift = 2.\allowbreak{}48}; \texttt{sim\_\allowbreak{}best = 0.\allowbreak{}815}; \texttt{COMET avg = 0.\allowbreak{}611}).}\par\smallskip
\begingroup
\small
\setlength{\tabcolsep}{3pt}
\begin{xltabular}{\textwidth}{@{}>{\RaggedRight\arraybackslash}p{0.24\textwidth}Y@{}}
\toprule
\textbf{Field} & \textbf{Text or label} \\
\midrule
\endfirsthead
\toprule
\textbf{Field} & \textbf{Text or label} \\
\midrule
\endhead
\midrule
\multicolumn{2}{r}{\footnotesize continued on next page}\\
\endfoot
\bottomrule
\endlastfoot
Passage & \texttt{sn4.\allowbreak{}25:\allowbreak{}2} \\
Pali source & \texttt{Kenāsi dummano tāta, purisaṁ kaṁ nu socasi; Mayaṁ taṁ rāgapāsena, āraññamiva kuñjaraṁ; Bandhitvā ānayissāma, vasago te bhavissati} \\
Sujato reference & "Why so downhearted, dad? What man are you upset about? We'll catch him with the snare of lust, like an elephant in the wild. We'll tie him up and bring him back--he'll fall under your sway!" \\
Thanissaro reference & "Why are you depressed, Daddy? For which man do you grieve? We--having bound him like a wilderness elephant with a snare of passion--will bring him to you. Under your power he'll be." \\
Bodhi reference & "Why are you despondent, father? Who's the man for whom you grieve? We'll catch him with the snare of lust As they catch the forest elephant. We'll bind him tightly and bring him back, And he'll be under your control." \\
Candidate translation & "Why are you dejected, dear, for which person do you grieve; we will bind you with the snare of lust, like an elephant in the forest, and bring you; you will be under our control." \\
Author adjudication & \texttt{MAJOR\_\allowbreak{}ERROR}; \texttt{AGENT\_\allowbreak{}OR\_\allowbreak{}ROLE\_\allowbreak{}ERROR}; high confidence \\
Primary judge panel & DeepSeek, Kimi, and Qwen all labeled \texttt{MAJOR\_\allowbreak{}ERROR}; all assigned \texttt{AGENT\_\allowbreak{}OR\_\allowbreak{}ROLE\_\allowbreak{}ERROR} with high confidence \\
\end{xltabular}
\endgroup

The issue is not register or style. The key sequence "Bandhitvā ānayissāma, vasago te bhavissati" is read by all three references as binding and bringing back "him" (the Buddha), so that he will come under Māra's control. The candidate changes the object to "you" and concludes "you will be under our control." This reverses the social relation in the verse: the addressee becomes the captured object rather than the beneficiary of the capture.

The complementary case below shows the opposite boundary: very high drift without error. In \texttt{sn1.\allowbreak{}20:\allowbreak{}3}, Venerable Samiddhi replies to a deity who has just urged him to enjoy sensual pleasures while young rather than let time pass him by. Samiddhi turns the time argument around: because the time of death or lost opportunity is unknown, he treats practice as urgent and does not postpone it until after enjoyment. The candidate is unusually compact and uses "beg" where the references prefer "seek/go for alms," but it preserves the verse's sequence and meaning.

\par\medskip\noindent\textbf{Table 14. Worked example of a high-drift valid variation (\texttt{normalized\_\allowbreak{}drift = 4.\allowbreak{}67}; \texttt{sim\_\allowbreak{}best = 0.\allowbreak{}887}; \texttt{COMET avg = 0.\allowbreak{}670}).}\par\smallskip
\begingroup
\small
\setlength{\tabcolsep}{3pt}
\begin{xltabular}{\textwidth}{@{}>{\RaggedRight\arraybackslash}p{0.24\textwidth}Y@{}}
\toprule
\textbf{Field} & \textbf{Text or label} \\
\midrule
\endfirsthead
\toprule
\textbf{Field} & \textbf{Text or label} \\
\midrule
\endhead
\midrule
\multicolumn{2}{r}{\footnotesize continued on next page}\\
\endfoot
\bottomrule
\endlastfoot
Passage & \texttt{sn1.\allowbreak{}20:\allowbreak{}3} \\
Pali source & \texttt{Kālaṁ vohaṁ na jānāmi, channo kālo na dissati; Tasmā abhutvā bhikkhāmi, mā maṁ kālo upaccagā} \\
Sujato reference & "I actually don't know the time; it's hidden and unseen. That's why I seek alms before eating, so that the time may not pass me by!" \\
Thanissaro reference & "I don't know my time. My time is hidden. It can't be seen. That's why, not having enjoyed, I go for alms: Don't let my time pass me by." \\
Bodhi reference & "I do not know what the time might be; The time is hidden and cannot be seen. Hence, without enjoying, I seek alms: Don't let the time pass me by!" \\
Candidate translation & "I do not know the time, the time is hidden and not seen; therefore I beg without having enjoyed -- let not the time pass me by." \\
Author adjudication & \texttt{VALID\_\allowbreak{}VARIATION}; high confidence \\
Primary judge panel & DeepSeek, Kimi, and Qwen all labeled \texttt{VALID\_\allowbreak{}VARIATION} with high confidence \\
\end{xltabular}
\endgroup

Here the high drift appears to be caused by phrasing rather than mistranslation. The candidate keeps all four semantic movements of the Pali: not knowing the time, the hiddenness of time, begging before eating or enjoyment, and the wish that time not pass by. The rendering "without having enjoyed" is close to Thanissaro's and Bodhi's wording, while "beg" is a more compressed equivalent for "bhikkhāmi" than "seek alms." This example is methodologically important because it prevents the audit from collapsing outlier status into error status. The same metric that successfully flags the major agent error in \texttt{sn4.\allowbreak{}25:\allowbreak{}2} also flags a defensible poetic variation here.

\subsection{Consistency Between Validation and Full Result Set}

The author-validation and full-result-set analyses agree on the main qualitative finding: higher drift is associated with a higher probability of major error, especially above \texttt{2.\allowbreak{}0}. In the author validation set, major errors rose from 2.0\% in \texttt{1.\allowbreak{}5-\allowbreak{}2.\allowbreak{}0} to 22.0\% above \texttt{2.\allowbreak{}0}. In the full judge-adjudicated result set, major errors rose from 7.9\% in \texttt{1.\allowbreak{}5-\allowbreak{}2.\allowbreak{}0} to 23.5\% above \texttt{2.\allowbreak{}0} when \texttt{2.\allowbreak{}0-\allowbreak{}3.\allowbreak{}0} and \texttt{\textgreater{}3.\allowbreak{}0} are combined.

The higher panel-labeled major-error rate in the full \texttt{1.\allowbreak{}5-\allowbreak{}2.\allowbreak{}0} band is partly explained by translator composition. The validation sample is approximately balanced by translator in each band, whereas the full \texttt{1.\allowbreak{}5-\allowbreak{}2.\allowbreak{}0} result set contains 219 Grok 4.3 instances and Grok has the highest major-error rate in that band (12.8\%). By contrast, GPT-5.5's full \texttt{1.\allowbreak{}5-\allowbreak{}2.\allowbreak{}0} panel-major rate is 2.1\%, close to the author-validation rate for the band. The robust conclusion is therefore not that \texttt{1.\allowbreak{}5-\allowbreak{}2.\allowbreak{}0} is clean, but that error composition in this band is model-dependent, while \texttt{\textgreater{}2.\allowbreak{}0}, and especially \texttt{\textgreater{}3.\allowbreak{}0}, is strongly enriched for major translation failures.

\section{Discussion}

This study separates three claims that are often collapsed in automatic translation evaluation. First, semantic distance from human references is not the same as error. Most \texttt{1.\allowbreak{}5-\allowbreak{}2.\allowbreak{}0} outliers, and a substantial fraction of even higher-drift outliers, were judged to be valid translation variations. This supports the multi-reference premise that legitimate translation variation must be modeled rather than treated as metric noise [Fomicheva, Specia, and Guzmán 2020]; [Wu, Wieting, and Smith 2024]. Second, semantic distance is still useful as a triage signal. Major errors cluster strongly in the upper tail, especially above \texttt{3.\allowbreak{}0}. Third, minor errors behave differently from major errors. They appear even at low and moderate drift and are harder for LLM judges to distinguish consistently from valid variation.

The practical implication is that multi-reference embedding outliers are best used as an audit queue, not as an automatic error label. This is consistent with the broader MT evaluation literature: automatic metrics are valuable, but their strongest role is often comparative or diagnostic rather than determinative for individual translation correctness [Post 2018]; [Mathur et al. 2020]; [Freitag et al. 2022]. For a classical language, this is useful, because it avoids the implausible claim that one English translation is the only correct rendering of a Pali passage, while still finding the region of the output distribution where serious translation failures are concentrated.

The model comparison also shows why adjudication is necessary. Gemini 3.1 Pro had the strongest automatic metrics, while GPT-5.5 had the lowest point estimate for adjudicated high-drift major-error burden. The confidence intervals for GPT-5.5, Gemini 3.1 Pro, and Claude Sonnet 4.6 overlap, so these three should not be treated as cleanly rank-ordered by major-error rate. Grok 4.3, by contrast, had both weaker automatic metrics and a much higher concentration of major errors in the high-drift tail. These findings would be blurred if the analysis reported only average embedding similarity, COMET, or raw outlier counts.

The positive result should not be missed. For the strongest models, most translations either stayed below the audit threshold or, when flagged, were often still judged defensible. This supports a practical use case for flagship LLMs as generators of first-draft translations, parallel renderings, or review candidates in a human scholarly workflow. Their value is speed, coverage, and the ability to produce plausible English renderings at scale. Their limitation is authority: they do not possess the lived practice context, tradition-specific accountability, or interpretive responsibility that shape a translator's final choices, and they still occasionally make major errors that a fluent surface can conceal.

For digital humanities (DH) and Buddhist-studies readers, the more important result is not that one model ranks ahead of another. It is that failures have different scholarly consequences. A mistranslated agent, reversed polarity, or doctrinally misleading term changes what a reader can responsibly infer from a canonical passage. The leaderboard is therefore best understood as an audit instrument: it identifies which systems produce more material requiring review and what kinds of review are needed. Translators, editors, teachers, students, and digital text projects could use such a workflow to prioritize human correction, not to certify machine translation as a replacement for philological work.

The error taxonomy points to recurring risk areas in Pali-to-English LLM translation and aligns with prior work showing that severe MT failures require more than average quality scores to detect [Raunak, Menezes, and Junczys-Dowmunt 2021]; [Guerreiro et al. 2023b]; [Dale et al. 2023]. Omission and truncation were the most common major failures, suggesting that completeness is a central audit target. Doctrinal terms were frequent across both minor and major errors, which is expected in Buddhist canonical translation: small lexical choices can have significant interpretive consequences. Findings on Ancient Greek translation are relevant here: rare technical terminology can be a strong predictor of catastrophic failure even when surrounding prose is translated well [Zainaldin et al. 2026]. Agent, role, negation, and logical-relation errors were less frequent but often severe because they directly change who acts, what is affirmed, or how clauses relate.

The shared-failure analysis suggests that high-drift major errors do not simply occur on the same hard passages for every model. Most observed high-drift major errors were single-model events. This matters for interpretation: a model-level error survey can reveal complementary failure modes that disappear in average score tables. Because below-threshold companion translations were not judged, this should be read as a result about drift-flagged major errors rather than definitive proof that the other models translated those passages correctly.

The broader methodological contribution is a reusable audit design for classical-to-modern translation workflows. Where multiple human references exist, reference spread can define a local envelope of acceptable variation; embedding drift can prioritize review; a small human-adjudicated validation set can calibrate the relation between drift and error; and an independent LLM judge panel can scale adjudication to the high-risk tail. The output is not just a leaderboard but a map of error burden, severity, and failure modes. This design is especially appropriate for settings where exhaustive expert review is too costly but unvalidated automatic scoring would be too blunt for scholarly use.

At the same time, the workflow makes a critical dependency visible. This study uses proprietary flagship models both as translation systems and as adjudication aids, while measuring them against human translations embedded in Buddhist textual traditions and contemporary digital infrastructures. That arrangement should not be mistaken for transferring textual authority to the models. The LLM judges operationalize a constrained task: compare a candidate with Pali and three references under a declared error taxonomy. They do not settle the meaning of a passage, adjudicate doctrinal interpretation, or replace the translator's responsibility to make situated choices for a particular audience. The value of the workflow is precisely that it keeps model output subordinate to human references, declared criteria, validation data, and inspectable examples.

Finally, AI translation of philosophical and religious texts is also ethically sensitive. Wider access is valuable, but accessibility should not be treated as separate from the communities, practitioners, translators, and scholars for whom these texts are living or authoritative traditions. Any practical deployment of AI-assisted translation in the religious domain should therefore include consultation with relevant communities and domain experts, especially when translations are intended for public religious, educational, or practical use.

\section{Limitations}

The main limitation is that the validation set is author-adjudicated rather than externally expert-adjudicated. The author adjudicator has formal training in Pali, but the labels are not an independent expert gold standard and do not include inter-rater reliability. This limitation is partly structural: because Pali is a low-resource philological language, it was not feasible for the authors to assemble a panel of validators who were both expert in Pali and sufficiently fluent in English translation judgment for this study. A future study could add multiple adjudicators to estimate intra-rater stability.

The study fully adjudicates only translations above \texttt{normalized\_\allowbreak{}drift \textgreater{} 1.\allowbreak{}5}. The validation set shows that minor errors also occur below this threshold, so total minor-error rates are likely underestimated by direct high-drift counts. Major errors below \texttt{1.\allowbreak{}5} were uncommon in the validation set, but not impossible. The analysis therefore does not treat below-threshold translations as error-free.

All translation is passage-level. Pali passages frequently contain ellipsis, repeated formulas, or dependencies on wider discourse context. This design is intentional and follows standard machine-translation evaluation practice, but it should be kept in mind when interpreting the results: the human reference translations were produced with access to broader textual context, whereas the LLMs translated only one passage at a time. For API efficiency, passages were batched by token count, but these batches were not designed to provide reliable discourse context because filtering and deduplication removed intervening passages and each passage was still requested as an independent JSON item.

The embedding triage model is Qwen3 Embedding. Qwen-family translator outputs were not included in the translator panel to avoid shared-family concerns. Even so, embedding choice can affect drift scores and should be treated as part of the measurement system and a design choice, rather than as a neutral oracle.

\section{Conclusion}

This study shows that flagship LLMs can be genuinely useful for Pali-to-English translation, especially as draft generators within a human-led workflow. It also shows why that usefulness must be paired with audit. Multi-reference semantic outliers are a strong starting point for review, but not a substitute for error adjudication. High normalized drift is a signal for major translation failures, especially in the extreme tail, while many moderate outliers remained valid variations and minor errors appeared across the drift spectrum. The strongest practical result is therefore methodological: reference-aware drift can prioritize review, but severity-labeled adjudication is needed to estimate real error burden.

Across four flagship translator models, serious failures were not evenly distributed. Grok 4.3 had the largest high-drift tail and the highest adjudicated major-error rate, while GPT-5.5, Claude Sonnet 4.6, and Gemini 3.1 Pro were closer and had overlapping confidence intervals for high-drift major errors. The most important error types were omission or truncation, doctrinal terminology, agent or role errors, additions or hallucinations, and negation or polarity errors. These are precisely the failures most likely to mislead readers of classical or doctrinal texts.

The proposed framework is intended to be reusable beyond Pali when multiple human references are available: define the local reference envelope, use embedding drift as a triage signal, calibrate with human adjudication, and scale review with an independent judge panel. The result is not only a model comparison, but an error survey that identifies where LLM translations fail, how severely they fail, and whether failures are shared across models or model-specific.

The immediate method introduced in this study requires existing reference translations, but its practical implication extends to texts that do not yet have translations in the target language. A reference-rich subset can be used as an audit bed to estimate which models are reliable enough for draft generation, which error types require human review, and which warning signs should trigger retranslation or escalation to a specialist. The resulting workflow does not authorize unsupervised publication of AI translations. It supports a more modest and useful practice: using LLMs to produce first drafts of untranslated material while understanding error modes, applying empirically grounded checks, targeted review priorities, and human judgment before any translation is treated as trustworthy.

\section{Acknowledgements}

The authors gratefully acknowledge Bhikkhu Sujato, Bhikkhu Bodhi, and Bhikkhu Thanissaro for their extraordinary work and sustained dedication in translating the texts of the Pali Canon and making them accessible to English readers. Their translations are not only acts of scholarship and service in their own right; they also made this study possible.

\section{Data and Code Availability}

A public reproducibility package is made available at \texttt{https:\allowbreak{}/\allowbreak{}/\allowbreak{}github.\allowbreak{}com/\allowbreak{}MateMetzger/\allowbreak{}pali-\allowbreak{}llm-\allowbreak{}error-\allowbreak{}audit}. The package includes the analysis and evaluation code, translation and adjudication prompts, corpus-construction metadata, passage and segment identifiers, text-length metadata, model identifiers, model-generated translation outputs, metric outputs, adjudication metadata, aggregate tables, and scripts needed to reproduce the analyses reported in this article.

The package does not redistribute the full aligned dataset file containing Pali source passages and aligned human reference translations, raw source files, aligned reference files, reference embeddings. The underlying source and reference translations are publicly available from the cited providers, but they are distributed under different terms: Sujato's SuttaCentral translations are released under Creative Commons Zero (CC0), Thanissaro's translations are available under Creative Commons Attribution-NonCommercial 4.0, and Bodhi's Wisdom Publications translations are available under Creative Commons Attribution-NonCommercial-NoDerivs 3.0 Unported. Because the latter license includes a NoDerivs condition, and because publishing aligned full-text extractions may constitute a derivative work, the public package will exclude all human translation references rather than treating the three translators differently. Researchers may reconstruct the aligned dataset locally for lawful scholarly use from publicly available material, subject to the applicable licenses, using the released passage manifest, prompts, pipeline scripts, and metadata.

\section{Generative AI Use Statement}

Generative artificial intelligence (AI) was used in two distinct ways in this study. First, the outputs of large language models are the object of research: the study evaluates LLM-generated Pali-to-English translations and uses independent LLM judges as part of the disclosed adjudication pipeline described in the Methods section. Second, GPT-5.5 and Claude 4.7 Opus were used by the authors as assistive tools for literature research, phrasing, language editing, code generation and formatting. Generative AI was not used as an independent author, to autonomously generate manuscript content, to independently analyze the data, or to derive the study's conclusions. All AI-assisted text, code, and analyses were reviewed, revised, verified, and accepted under author control. The authors take responsibility for the accuracy, integrity, and final content of the submission.

\section{Conflict of Interest and Funding}

The authors declare no conflicts of interest. This study received no external funding and was self-funded by the authors.

\renewcommand{\refname}{Works Cited}

\appendix

\section{Translation Prompt}

\subsection{Translation System Prompt}

\begin{CodeBlock}

Translate the following Pali passages into English.

Return a JSON object with the same keys and the English translations as values.

Do not add text that is not present in the input.



IMPORTANT: For valid JSON, never use double quotes inside translation values.

- Use single quotes for dialogue: 'Hello,' he said.

- Use backticks for nested quotes: 'He said `hello` to me.'



Example:

Input: {"mn1:5": "Pathavī pathavīti sañjānāti.", "mn1:6": "Āpo āpoti sañjānāti."}

Output: {"mn1:5": "They perceive earth as earth.", "mn1:6": "They perceive water as water."}



Output ONLY the JSON object, no other text.

\end{CodeBlock}

\subsection{Translation User Prompt}

The user message is a JSON object mapping passage IDs to Pali source text:

\begin{CodeBlock}

{

  "mn1:5": "Pathavī pathavīti sañjānāti.",

  "mn1:6": "Āpo āpoti sañjānāti."

}

\end{CodeBlock}

The model must return a JSON object with exactly the same keys and English translations as values.

\section{LLM Judge Prompt}

\subsection{Judge System Prompt}

\begin{CodeBlock}

You are an expert adjudicator for Pali-to-English translation quality.



You will be shown a Pali source passage, three established English reference translations, and one candidate translation.



Important: the candidate has been selected for translation-quality adjudication. It may be an embedding outlier, near-outlier, or control example. Your task is to determine whether the candidate is a legitimate translation variation or contains a translation issue.



Do not assume that selection means error. A translation may be correct even if it differs from all three references in register, terminology, syntax, compression, expansion, or interpretive style.



Mark MAJOR_ERROR only when the candidate appears to mistranslate, omit, add, distort, or leave unresolved content from the Pali source in a way that materially changes or obscures the meaning. If you are unsure because the passage lacks wider discourse context, use UNSURE_CONTEXT_NEEDED.



Operational boundary:

- Use MINOR_ERROR for a real, correction-worthy translation problem that does not substantially change the main meaning. The core agents, actions, negation, doctrinal point, logical relation, and practical implication remain recoverable.

- Use MAJOR_ERROR when the candidate materially changes, obscures, or loses the meaning. A reader relying only on the candidate translation would be materially misled.

- When in doubt between minor and major error, prefer MINOR_ERROR unless the expected misunderstanding is material.



Special policy for Pali ellipsis:

- If the source contains …pe… and the candidate leaves it unresolved but the English remains readable and the main meaning is recoverable, use label MINOR_ERROR and error_category UNRESOLVED_SOURCE_ELLIPSIS.

- If unresolved ellipsis leaves a dangling fragment, omits the main predicate, or makes the intended meaning unrecoverable, use label MAJOR_ERROR and error_category UNRESOLVED_SOURCE_ELLIPSIS.

- Do not penalize a candidate merely for being more concise than the references if the source itself is abbreviated and the meaning remains recoverable.



The label field must be exactly one of: VALID_VARIATION, MINOR_ERROR, MAJOR_ERROR, UNSURE_CONTEXT_NEEDED, REFERENCE_OR_ALIGNMENT_PROBLEM.

Never put an error category in the label field. Error categories belong only in error_category.



Return only valid JSON. Do not use Markdown.

\end{CodeBlock}

\subsection{Judge User Prompt Template}

\begin{CodeBlock}

Adjudicate this Pali-to-English candidate translation.



Allowed top-level labels:

- VALID_VARIATION: defensible translation; differences mainly reflect style, register, terminology, syntax, compression, expansion, or interpretive variation.

- MINOR_ERROR: a real, correction-worthy translation problem that does not substantially change the passage's main meaning. The core agents, actions, negation, doctrinal point, logical relation, and practical implication remain recoverable.

- MAJOR_ERROR: a translation problem that materially changes, obscures, or loses the passage's meaning. A reader relying only on the candidate translation would be materially misled.

- UNSURE_CONTEXT_NEEDED: cannot fairly adjudicate without wider sutta context.

- REFERENCE_OR_ALIGNMENT_PROBLEM: the references or passage alignment appear to be the main source of the discrepancy.



The label field must be exactly one of those five labels. Never put an error category such as UNRESOLVED_SOURCE_ELLIPSIS, OMISSION_OR_TRUNCATION, or DOCTRINAL_TERM_ERROR in the label field.



Minor vs major error decision rule:

- If the candidate would let a reader understand the main point, but a careful translator would still correct a local mistake, use MINOR_ERROR.

- If the candidate would cause a reader to misunderstand who did what, whether something was affirmed or denied, what was included in a list or number, what relation holds between clauses, or what doctrinal point is being made, use MAJOR_ERROR.

- When in doubt between MINOR_ERROR and MAJOR_ERROR, prefer MINOR_ERROR unless the expected misunderstanding is material.



If label is MAJOR_ERROR or MINOR_ERROR, choose one primary error_category:

- OMISSION_OR_TRUNCATION

- ADDITION_OR_HALLUCINATION

- NEGATION_OR_POLARITY_ERROR

- AGENT_OR_ROLE_ERROR

- DOCTRINAL_TERM_ERROR

- LIST_OR_NUMBER_ERROR

- UNRESOLVED_SOURCE_ELLIPSIS

- WRONG_RELATION_OR_LOGIC

- OVERLY_FREE_PARAPHRASE

- GRAMMAR_OR_FLUENCY_PROBLEM

- OTHER



If label is VALID_VARIATION, UNSURE_CONTEXT_NEEDED, or REFERENCE_OR_ALIGNMENT_PROBLEM, use error_category: NONE.



For unresolved Pali ellipsis:

- recoverable/readable unresolved ellipsis: label MINOR_ERROR, error_category UNRESOLVED_SOURCE_ELLIPSIS.

- unrecoverable/dangling unresolved ellipsis: label MAJOR_ERROR, error_category UNRESOLVED_SOURCE_ELLIPSIS.



Use severity: none, low, medium, or high.

Use confidence: low, medium, or high.



Return exactly this JSON shape, replacing the example values as appropriate:

{

  "label": "...",

  "error_category": "...",

  "severity": "...",

  "confidence": "...",

  "is_major_error": false,

  "is_valid_variation": true,

  "cause_summary": "one sentence",

  "evidence": "brief explanation grounded in the Pali and references",

  "problem_phrase": "exact problematic candidate phrase, or empty string",

  "suggested_correction": "brief correction if clear, or empty string"

}



Passage ID: {passage_id}

Normalized drift: {normalized_drift}

Nearest-reference similarity: {sim_best}



Pali source:

{pali}



Sujato reference:

{sujato}



Thanissaro reference:

{thanissaro}



Bodhi reference:

{bodhi}



Candidate translation:

{candidate}

\end{CodeBlock}

\section{Alignment Prompt}

\begin{CodeBlock}

ROLE

You are an alignment engine. Your task is to extract English text from an existing translation to match specific Pāli segment IDs.



INPUT

For each segment ID you are given:

- the Pāli text (semantic anchor - this determines what content belongs to the segment)

- Sujato's English translation (segmentation reference only)



You are also given the FULL English translation by the target translator.



GOAL

For every segment ID, extract the corresponding English text from the translator that expresses the same meaning as the Pāli.



IMPORTANT RULES



1) Pāli is the authority.

   Use the Pāli text to decide what content belongs to each segment.



2) Sujato is only a guide.

   His segmentation helps you understand boundaries, but the target translator may split/merge differently.



3) MATCH THE PALI STRUCTURE.

   The OUTPUT must mirror the structure of the PALI segment:



   a) If the Pali contains "…" or is abbreviated:

      → Extract ONLY the corresponding term(s) from the translator

      → Do NOT expand to the full sentence

      → Example: Pali "viññātaṁ …" + Sujato "the known …"

        → Output just "the cognized" (not the full paragraph)



   b) If the Pali is fully spelled out (no ellipsis):

      → Extract the complete matching text from the translator

      → If the translator abbreviates with "..." but Pali is full,

        expand by finding the earlier template and substituting the terms



4) Minimal faithful extraction.

   Extract the smallest text that expresses the Pāli meaning.

   Prefer contiguous substrings from the original.



5) Respect text order.

   Process the translation in order. Don't reuse non-repetitive text.



6) Filter noise.

   Discard footnotes, section headers (unless part of translation), editor notes, bracketed references.



7) Null policy.

   Output null ONLY if you genuinely cannot find matching content.

   Remember: abbreviated Pali → short output (just the term).



OUTPUT FORMAT (STRICT JSON)

Return valid JSON with exactly the same keys as the input, in the same order.

Each key maps to either a string (extracted text) or null.



Example:

{

  "mn1:3.1": "Here, monks, an untaught ordinary person...",

  "mn1:3.2": "He perceives earth as earth.",

  "mn1:3.3": null

}



Do NOT include Pāli, Sujato, or any extra fields.

Do NOT add commentary or explanation.

JSON only.

\end{CodeBlock}

\section{Author Validation Instructions}

The author validation subset uses the same top-level label schema as the LLM judges, but with a lighter annotation form to keep the human workload realistic. The author adjudicator sees the Pali source, three references, candidate translation, normalized drift score, and metric metadata, but does not see candidate model identity or LLM judge labels.

Author adjudication fields:

\begin{CodeBlock}

{

  "label": "...",

  "error_category": "...",

  "severity": "...",

  "confidence": "...",

  "notes": ""

}

\end{CodeBlock}

Required fields:

\begin{itemize}
\item \texttt{label}: required for every item.
\item \texttt{error\_\allowbreak{}category}: required for \texttt{MINOR\_\allowbreak{}ERROR} and \texttt{MAJOR\_\allowbreak{}ERROR}; use \texttt{NONE} otherwise.
\item \texttt{severity}: required for every item.
\item \texttt{confidence}: recommended for every item.
\item \texttt{notes}: optional; use when helpful, especially for errors, uncertainty, or possible reference/alignment problems.
\end{itemize}

Author label guidance:

\begin{itemize}
\item Use \texttt{VALID\_\allowbreak{}VARIATION} when the candidate is defensible even if it differs from all references.
\item Use \texttt{MINOR\_\allowbreak{}ERROR} for a real, correction-worthy translation problem that does not substantially change the passage's main meaning. The core agents, actions, negation, doctrinal point, logical relation, and practical implication remain recoverable.
\item Use \texttt{MAJOR\_\allowbreak{}ERROR} only when the candidate materially changes, obscures, or loses the passage's meaning. A reader relying only on the candidate translation would be materially misled.
\item Use \texttt{REFERENCE\_\allowbreak{}OR\_\allowbreak{}ALIGNMENT\_\allowbreak{}PROBLEM} when the item itself appears problematic.
\item Use \texttt{UNSURE\_\allowbreak{}CONTEXT\_\allowbreak{}NEEDED} when wider sutta context is necessary for a fair decision.
\item When in doubt between \texttt{MINOR\_\allowbreak{}ERROR} and \texttt{MAJOR\_\allowbreak{}ERROR}, prefer \texttt{MINOR\_\allowbreak{}ERROR} unless the expected misunderstanding is material.
\end{itemize}

The author should make a best effort to ground each decision in the Pali source and the three references, while acknowledging uncertainty where relevant.

\section{Supplementary Tables}

\par\medskip\noindent\textbf{Appendix Table E1. Majority aggregation rule for the three-judge panel.}\par\smallskip
\begingroup
\small
\setlength{\tabcolsep}{3pt}
\begin{xltabular}{\textwidth}{@{}>{\RaggedRight\arraybackslash}p{0.24\textwidth}Y@{}}
\toprule
\textbf{Panel condition} & \textbf{Aggregate label} \\
\midrule
\endfirsthead
\toprule
\textbf{Panel condition} & \textbf{Aggregate label} \\
\midrule
\endhead
\midrule
\multicolumn{2}{r}{\footnotesize continued on next page}\\
\endfoot
\bottomrule
\endlastfoot
At least 2 of 3 judges label \texttt{MAJOR\_\allowbreak{}ERROR} & \texttt{MAJOR\_\allowbreak{}ERROR} \\
At least 2 of 3 judges label \texttt{MINOR\_\allowbreak{}ERROR} or \texttt{MAJOR\_\allowbreak{}ERROR}, but fewer than 2 label \texttt{MAJOR\_\allowbreak{}ERROR} & \texttt{MINOR\_\allowbreak{}ERROR} \\
At least 2 of 3 judges label \texttt{VALID\_\allowbreak{}VARIATION} & \texttt{VALID\_\allowbreak{}VARIATION} \\
Otherwise & \texttt{UNCERTAIN} \\
\end{xltabular}
\endgroup

\par\medskip\noindent\textbf{Appendix Table E2. Size of the high-drift result set by translator.}\par\smallskip
\begingroup
\small
\setlength{\tabcolsep}{3pt}
\begin{xltabular}{\textwidth}{@{}>{\RaggedRight\arraybackslash}p{0.24\textwidth}Y@{}}
\toprule
\textbf{Translator} & \textbf{\texttt{\textgreater{}1.\allowbreak{}5} instances} \\
\midrule
\endfirsthead
\toprule
\textbf{Translator} & \textbf{\texttt{\textgreater{}1.\allowbreak{}5} instances} \\
\midrule
\endhead
\midrule
\multicolumn{2}{r}{\footnotesize continued on next page}\\
\endfoot
\bottomrule
\endlastfoot
GPT-5.5 & 224 \\
Claude Sonnet 4.6 & 302 \\
Gemini 3.1 Pro & 257 \\
Grok 4.3 & 420 \\
\textbf{Total} & \textbf{1,203} \\
\end{xltabular}
\endgroup

\par\medskip\noindent\textbf{Appendix Table E3. Author-adjudicated validation labels.}\par\smallskip
\begingroup
\small
\setlength{\tabcolsep}{3pt}
\begin{xltabular}{\textwidth}{@{}>{\RaggedRight\arraybackslash}p{0.22\textwidth}Y Y@{}}
\toprule
\textbf{Author label} & \textbf{Count} & \textbf{Percent} \\
\midrule
\endfirsthead
\toprule
\textbf{Author label} & \textbf{Count} & \textbf{Percent} \\
\midrule
\endhead
\midrule
\multicolumn{3}{r}{\footnotesize continued on next page}\\
\endfoot
\bottomrule
\endlastfoot
\texttt{VALID\_\allowbreak{}VARIATION} & 227 & 75.7\% \\
\texttt{MINOR\_\allowbreak{}ERROR} & 58 & 19.3\% \\
\texttt{MAJOR\_\allowbreak{}ERROR} & 15 & 5.0\% \\
\end{xltabular}
\endgroup

\par\medskip\noindent\textbf{Appendix Table E4. Confusion matrix between author labels and majority-panel labels.}\par\smallskip
\begingroup
\small
\setlength{\tabcolsep}{3pt}
\begin{xltabular}{\textwidth}{@{}>{\RaggedRight\arraybackslash}p{0.24\textwidth}Y Y Y@{}}
\toprule
\textbf{Author  Panel} & \textbf{Valid variation} & \textbf{Minor error} & \textbf{Major error} \\
\midrule
\endfirsthead
\toprule
\textbf{Author  Panel} & \textbf{Valid variation} & \textbf{Minor error} & \textbf{Major error} \\
\midrule
\endhead
\midrule
\multicolumn{4}{r}{\footnotesize continued on next page}\\
\endfoot
\bottomrule
\endlastfoot
Valid variation & 213 & 14 & 0 \\
Minor error & 36 & 17 & 5 \\
Major error & 0 & 5 & 10 \\
\end{xltabular}
\endgroup

\par\medskip\noindent\textbf{Appendix Table E5. Direct observed high-drift error counts expressed as lower-bound percentages of the full 1,700-passage dataset.}\par\smallskip
\begingroup
\scriptsize
\setlength{\tabcolsep}{2pt}
\begin{xltabular}{\textwidth}{@{}Y Y Y Y Y Y@{}}
\toprule
\textbf{Translator} & \textbf{High-drift instances} & \textbf{Minor errors} & \textbf{Major errors} & \textbf{Major 95\% confidence interval} & \textbf{Any high-drift error} \\
\midrule
\endfirsthead
\toprule
\textbf{Translator} & \textbf{High-drift instances} & \textbf{Minor errors} & \textbf{Major errors} & \textbf{Major 95\% confidence interval} & \textbf{Any high-drift error} \\
\midrule
\endhead
\midrule
\multicolumn{6}{r}{\footnotesize continued on next page}\\
\endfoot
\bottomrule
\endlastfoot
GPT-5.5 & 224 & 1.06\% & 0.53\% & 0.28-1.00\% & 1.59\% \\
Claude Sonnet 4.6 & 302 & 2.47\% & 1.53\% & 1.05-2.23\% & 4.00\% \\
Gemini 3.1 Pro & 257 & 2.12\% & 1.06\% & 0.67-1.67\% & 3.18\% \\
Grok 4.3 & 420 & 4.29\% & 6.82\% & 5.72-8.12\% & 11.12\% \\
\end{xltabular}
\endgroup

\par\medskip\noindent\textbf{Appendix Table E6. Majority-panel minor-error categories in the high-drift result set.}\par\smallskip
\begingroup
\small
\setlength{\tabcolsep}{3pt}
\begin{xltabular}{\textwidth}{@{}>{\RaggedRight\arraybackslash}p{0.24\textwidth}Y@{}}
\toprule
\textbf{Minor-error category} & \textbf{Count} \\
\midrule
\endfirsthead
\toprule
\textbf{Minor-error category} & \textbf{Count} \\
\midrule
\endhead
\midrule
\multicolumn{2}{r}{\footnotesize continued on next page}\\
\endfoot
\bottomrule
\endlastfoot
\texttt{UNRESOLVED\_\allowbreak{}SOURCE\_\allowbreak{}ELLIPSIS} & 60 \\
\texttt{DOCTRINAL\_\allowbreak{}TERM\_\allowbreak{}ERROR} & 50 \\
\texttt{ADDITION\_\allowbreak{}OR\_\allowbreak{}HALLUCINATION} & 17 \\
\texttt{OMISSION\_\allowbreak{}OR\_\allowbreak{}TRUNCATION} & 10 \\
\texttt{AGENT\_\allowbreak{}OR\_\allowbreak{}ROLE\_\allowbreak{}ERROR} & 9 \\
\texttt{GRAMMAR\_\allowbreak{}OR\_\allowbreak{}FLUENCY\_\allowbreak{}PROBLEM} & 7 \\
\texttt{LIST\_\allowbreak{}OR\_\allowbreak{}NUMBER\_\allowbreak{}ERROR} & 5 \\
\texttt{OTHER} & 4 \\
\texttt{WRONG\_\allowbreak{}RELATION\_\allowbreak{}OR\_\allowbreak{}LOGIC} & 3 \\
\texttt{NEGATION\_\allowbreak{}OR\_\allowbreak{}POLARITY\_\allowbreak{}ERROR} & 2 \\
\texttt{OVERLY\_\allowbreak{}FREE\_\allowbreak{}PARAPHRASE} & 2 \\
\end{xltabular}
\endgroup

\par\medskip\noindent\textbf{Appendix Table E7. Passage-level overlap of any errors across translator models.}\par\smallskip
\begingroup
\small
\setlength{\tabcolsep}{3pt}
\begin{xltabular}{\textwidth}{@{}>{\RaggedRight\arraybackslash}p{0.24\textwidth}Y@{}}
\toprule
\textbf{Number of translator models with any error on the same passage} & \textbf{Passage count} \\
\midrule
\endfirsthead
\toprule
\textbf{Number of translator models with any error on the same passage} & \textbf{Passage count} \\
\midrule
\endhead
\midrule
\multicolumn{2}{r}{\footnotesize continued on next page}\\
\endfoot
\bottomrule
\endlastfoot
1 & 195 \\
2 & 47 \\
3 & 15 \\
4 & 1 \\
\end{xltabular}
\endgroup

\par\medskip\noindent\textbf{Appendix Table E8. Illustrative shared-major-error passages and their failure profiles across translator models.}\par\smallskip
\begingroup
\small
\setlength{\tabcolsep}{3pt}
\begin{xltabular}{\textwidth}{@{}>{\RaggedRight\arraybackslash}p{0.24\textwidth}Y@{}}
\toprule
\textbf{Item} & \textbf{Text} \\
\midrule
\endfirsthead
\toprule
\textbf{Item} & \textbf{Text} \\
\midrule
\endhead
\midrule
\multicolumn{2}{r}{\footnotesize continued on next page}\\
\endfoot
\bottomrule
\endlastfoot
\textbf{Passage} & \texttt{sn4.\allowbreak{}24:\allowbreak{}10}: 3 major, 0 minor, 0 valid, 1 below threshold / not judged \\
\textbf{Pali} & \texttt{Aladdhā tattha assādaṁ, vāyasetto apakkame; Kākova selamāsajja, nibbijjāpema gotamā} \\
\textbf{Sujato reference} & "But finding nothing tasty, the crow left that place. Like the crow that pecked the stone, I leave Gotama disappointed." \\
\textbf{Thanissaro reference} & "but not getting anything tasty there, the crow went away. Like a crow attacking the rock, I weary myself with Gotama." \\
\textbf{Bodhi reference} & "But because he found nothing tasty there, The crow departed from that spot. Just like the crow that attacked the stone, We leave Gotama disappointed." \\
\textbf{GPT-5.5} & [\texttt{MAJOR\_\allowbreak{}ERROR}] "Not finding any enjoyment there, the crow's offspring went away. Like a crow that has attacked a rock, we leave Gotama in disappointment." \\
\textbf{Claude} & [\texttt{MAJOR\_\allowbreak{}ERROR}] "Having found no delight there, let the best of crows depart; like a crow that has struck a rock, let us leave in disappointment, Gotama." \\
\textbf{Gemini} & [below threshold / not judged] "Finding nothing tasty there, the crow left. Like a crow pecking at a rock, I leave Gotama disappointed." \\
\textbf{Grok} & [\texttt{MAJOR\_\allowbreak{}ERROR}] "Not having found delight there, the wind-swept one departs; like a crow striking a rock, we are wearied of Gotama." \\
\textbf{Main failure pattern} & The only three-major passage. Claude, GPT-5.5, and Grok all stumbled over the crow simile and agent structure. GPT-5.5 introduced "crow's offspring", Claude shifted the verse toward an exhortation, and Grok produced a nonsensical agent. \\
\addlinespace
\textbf{Passage} & \texttt{an4.\allowbreak{}24:\allowbreak{}5}: 2 major, 2 minor, 0 valid, 0 below threshold / not judged \\
\textbf{Pali} & \texttt{Yaṁ, bhikkhave .\allowbreak{}.\allowbreak{}.\allowbreak{}pe.\allowbreak{}.\allowbreak{}.\allowbreak{} tamahaṁ jānāmi ca na ca jānāmīti vadeyyaṁ} \\
\textbf{Sujato reference} & "If I were to say that 'I both know and do not know ... the world with its gods', that would be just the same." \\
\textbf{Thanissaro reference} & "If I were to say, 'I both know and don't know whatever in this world ... is seen, heard, sensed, cognized ... pondered by the intellect,' that would be just the same." \\
\textbf{Bodhi reference} & "if I were to say, 'In this world with its devas ... whatever is seen, heard, sensed, cognized ... that I both know and do not know,' that too would be just the same." \\
\textbf{GPT-5.5} & [\texttt{MINOR\_\allowbreak{}ERROR}] "Mendicants, if I were to say of what ... 'I know it and I do not know it,' see, that would be just the same." \\
\textbf{Claude} & [\texttt{MAJOR\_\allowbreak{}ERROR}] "Monks, if I were to say of that which I know ... 'I both know it and do not know it,' that too would be just the same." \\
\textbf{Gemini} & [\texttt{MINOR\_\allowbreak{}ERROR}] "If I were to say: 'I both know and do not know ...' that would be just the same." \\
\textbf{Grok} & [\texttt{MAJOR\_\allowbreak{}ERROR}] "Whatever, monks ...pe... that I would say 'I know and I do not know', that would be just the same for him." \\
\textbf{Main failure pattern} & The hardest four-model case by total error count. All four translators mishandled unresolved source ellipsis; Claude and Grok were judged major because the referent or logical structure became unrecoverable, while GPT-5.5 and Gemini were judged minor. \\
\addlinespace
\textbf{Passage} & \texttt{sn4.\allowbreak{}25:\allowbreak{}2}: 2 major, 1 minor, 0 valid, 1 below threshold / not judged \\
\textbf{Pali} & \texttt{Bandhitvā ānayissāma, vasago te bhavissati} \\
\textbf{Sujato reference} & "We'll tie him up and bring him back--he'll fall under your sway!" \\
\textbf{Thanissaro reference} & "We--having bound him like a wilderness elephant with a snare of passion--will bring him to you. Under your power he'll be." \\
\textbf{Bodhi reference} & "We'll bind him tightly and bring him back, And he'll be under your control." \\
\textbf{GPT-5.5} & [below threshold / not judged] "We shall bind him with the snare of passion, like a forest elephant, and bring him here; he will be under your control." \\
\textbf{Claude} & [\texttt{MAJOR\_\allowbreak{}ERROR}] "We will bind you with the snare of passion, like an elephant in the forest, and bring you back; you will be under our control." \\
\textbf{Gemini} & [\texttt{MINOR\_\allowbreak{}ERROR}] "We'll catch him in the snare of lust, and lead him like an elephant from the forest. He'll be under your sway!" \\
\textbf{Grok} & [\texttt{MAJOR\_\allowbreak{}ERROR}] "we will bind you with the snare of lust, like an elephant in the forest, and bring you; you will be under our control." \\
\textbf{Main failure pattern} & Claude and Grok shifted the object of binding from the target man to the addressee, reversing who would be captured and controlled. Gemini had a minor omission/truncation issue. \\
\addlinespace
\textbf{Passage} & \texttt{sn11.\allowbreak{}22:\allowbreak{}5}: 2 major, 0 minor, 2 valid, 0 below threshold / not judged \\
\textbf{Pali} & \texttt{Kuddhāhaṁ na pharusaṁ brūmi, Na ca dhammāni kittaye; Sanniggaṇhāmi attānaṁ, Sampassaṁ atthamattano} \\
\textbf{Sujato reference} & "When I do get angry I don't speak harshly, nor do I advertise my own virtues. I carefully restrain myself out of regard for my own welfare." \\
\textbf{Thanissaro reference} & "When angry, I don't speak harshly or insist on my virtues. I keep myself well under control with an eye to my own good." \\
\textbf{Bodhi reference} & "When I'm angry I don't speak harshly And I don't praise my virtues. I keep myself well restrained Out of regard for my own good." \\
\textbf{GPT-5.5} & [\texttt{VALID\_\allowbreak{}VARIATION}] "When angry, I do not speak harshly, nor do I praise my virtues. I restrain myself, seeing my own good." \\
\textbf{Claude} & [\texttt{MAJOR\_\allowbreak{}ERROR}] "When angry, I do not speak harshly, nor do I praise unworthy things; I restrain myself, seeing my own welfare." \\
\textbf{Gemini} & [\texttt{VALID\_\allowbreak{}VARIATION}] "When angry I don't speak harshly, nor do I praise my own virtues. I keep myself in check, seeing my own true goal." \\
\textbf{Grok} & [\texttt{MAJOR\_\allowbreak{}ERROR}] "When angry, I do not speak harshly, nor do I praise unrighteous things; I restrain myself, seeing my own welfare." \\
\textbf{Main failure pattern} & Claude and Grok rendered \texttt{dhammāni} as "unworthy/unrighteous things", reversing the sense from not advertising virtues or qualities. GPT-5.5 and Gemini were judged valid, showing that this was not an unavoidable passage-level failure. \\
\addlinespace
\textbf{Passage} & \texttt{sn46.\allowbreak{}53:\allowbreak{}5}: 2 major, 0 minor, 2 valid, 0 below threshold / not judged \\
\textbf{Pali} & \texttt{Seyyathāpi .\allowbreak{}.\allowbreak{}.\allowbreak{} parittaṁ aggiṁ ujjāletukāmo .\allowbreak{}.\allowbreak{}.\allowbreak{} allāni .\allowbreak{}.\allowbreak{}.\allowbreak{} udakavātañca dadeyya, paṁsukena ca okireyya .\allowbreak{}.\allowbreak{}.\allowbreak{} No hetaṁ, bhante} \\
\textbf{Sujato reference} & "Suppose someone wanted to make a small fire flare up. If they toss wet grass, cow-dung, and timber on it, spray it with water, and scatter dirt on it, could they make it flare up? No, sir." \\
\textbf{Thanissaro reference} & "Just as if a man, wanting to make a small fire blaze up ... were to give it a spray of water and smother it with dust. Is it possible that he would make the small fire blaze up? No, lord." \\
\textbf{Bodhi reference} & "Suppose ... a man wants to make a small fire flare up. If he throws wet grass, wet cowdung, and wet timber into it, sprays it with water, and scatters soil over it, would he be able to make that small fire flare up? No, venerable sir." \\
\textbf{GPT-5.5} & [\texttt{MAJOR\_\allowbreak{}ERROR}] "Suppose ... a man wanted to kindle a small fire. He would put wet grass on it ... give it water and wind, and scatter dust over it. Would that man be able to kindle the small fire? No, venerable sir." \\
\textbf{Claude} & [\texttt{VALID\_\allowbreak{}VARIATION}] "Monks, suppose a person wanted to kindle a small fire. If they were to put wet grass on it, wet cow dung, and wet wood, and were to give it a spray of water and cover it with dust ... No, Venerable Sir." \\
\textbf{Gemini} & [\texttt{VALID\_\allowbreak{}VARIATION}] "Suppose a person wanted to light a small fire. They'd add wet grass, wet cow-dung, and wet timber; they'd spray it with water and scatter dirt on it. Could they light a small fire? No, sir." \\
\textbf{Grok} & [\texttt{MAJOR\_\allowbreak{}ERROR}] "Suppose ... a person wants to kindle a small fire. If he were to throw wet grass, wet cow dung, and wet sticks on it, and blow on it with water and cover it with dust ... No, venerable sir." \\
\textbf{Main failure pattern} & GPT-5.5 and Grok distorted the water-spray action in the fire simile by adding or mishandling "wind"; Claude and Gemini were judged valid. \\
\addlinespace
\textbf{Passage} & \texttt{sn5.\allowbreak{}3:\allowbreak{}5}: 2 major, 0 minor, 2 valid, 0 below threshold / not judged \\
\textbf{Pali} & \texttt{Accantaṁ mataputtāmhi, purisā etadantikā; Na socāmi na rodāmi, na taṁ bhāyāmi āvuso} \\
\textbf{Sujato reference} & "I've got over the death of children, and I'm finished with men. I don't grieve or lament, and I'm not afraid of you, sir!" \\
\textbf{Thanissaro reference} & "I've gotten past the killing of sons, have made that the end to (my search for) men. I don't grieve, I don't weep--and I'm not afraid of you, my friend." \\
\textbf{Bodhi reference} & "I've gotten past the death of sons; With this, the search for men has ended. I do not sorrow, I do not weep, Nor do I fear you, friend." \\
\textbf{GPT-5.5} & [\texttt{MAJOR\_\allowbreak{}ERROR}] "I am one whose son is utterly dead; men too have come to an end in this way. I do not sorrow, I do not weep, I am not afraid of you, friend." \\
\textbf{Claude} & [\texttt{VALID\_\allowbreak{}VARIATION}] "I have utterly done with dead children; men have come to an end for me. I do not grieve, I do not weep, and I do not fear you, friend." \\
\textbf{Gemini} & [\texttt{VALID\_\allowbreak{}VARIATION}] "I've completely gotten over the death of my children, and I'm done with men. I don't sorrow or weep, and I'm not afraid of you, sir!" \\
\textbf{Grok} & [\texttt{MAJOR\_\allowbreak{}ERROR}] "I am one whose child has utterly died; men have that as their end. I do not sorrow, I do not weep, I do not fear you, friend." \\
\textbf{Main failure pattern} & GPT-5.5 and Grok turned a verse about having gone beyond grief into a more literal or nihilistic claim about death and men, while Claude and Gemini were judged valid. \\
\end{xltabular}
\endgroup
\end{document}